\documentclass[10pt,journal,compsoc]{IEEEtran}
\usepackage{times}
\usepackage{epsfig}
\usepackage{graphicx}
\usepackage{amsmath}
\usepackage{amssymb}
\usepackage{graphicx}
\usepackage[linesnumbered,boxed,ruled,commentsnumbered]{algorithm2e}
\usepackage{booktabs}
\usepackage{threeparttable}
\usepackage{multirow}
\usepackage{array}
\usepackage{color}
\usepackage{footmisc}
\usepackage{url}
\usepackage{ragged2e}
\usepackage[colorlinks,
linkcolor=black,
anchorcolor=black,
citecolor=black
]{hyperref}
\begin{document}

\title{DUT-LFSaliency: Versatile Dataset and Light Field-to-RGB Saliency Detection}
\author{
	Yongri~Piao,
        Zhengkun~Rong,
        Shuang~Xu,
        Miao~Zhang, and~Huchuan~Lu
\IEEEcompsocitemizethanks{\IEEEcompsocthanksitem Y.R. Piao, Z.K. Rong and H.C. Lu are with the School of Information and Communication Engineering, Dalian University of Technology, China
\IEEEcompsocthanksitem M. Zhang and S. Xu are with the International School of Information and Software Engineering, Dalian University of Technology, China}}

%
\markboth{IEEE TRANSACTIONS ON PATTERN ANALYSIS AND MACHINE INTELLIGENCE}%
{Shell \MakeLowercase{\textit{et al.}}: Bare Demo of IEEEtran.cls for Computer Society Journals}

\IEEEtitleabstractindextext{%
	\begin{abstract}
		\justifying
		Light field data exhibit favorable characteristics conducive to saliency detection. The success of learning-based light field saliency detection is heavily dependent on how a comprehensive dataset can be constructed for higher generalizability of models, how high dimensional light field data can be effectively exploited, and how a flexible model can be designed to achieve versatility for desktop computers and mobile devices. To answer these questions, first we introduce a large-scale dataset to enable versatile applications for RGB, RGB-D and light field saliency detection, containing 102 classes and 4204 samples. Second, we present an asymmetrical two-stream model consisting of the Focal stream and RGB stream. The Focal stream is designed to achieve higher performance on desktop computers and transfer focusness knowledge to the RGB stream, relying on two tailor-made modules. The RGB stream guarantees the flexibility and memory/computation efficiency on mobile devices through three distillation schemes. Experiments demonstrate that our Focal stream achieves state-of-the-arts performance. The RGB stream achieves Top-2 F-measure on DUTLF-V2, which tremendously minimizes the model size by 83\% and boosts FPS by 5 times, compared with the best performing method. Furthermore, our proposed distillation schemes are applicable to RGB saliency models, achieving impressive performance gains while ensuring flexibility.
	\end{abstract}
	

}

\maketitle

\IEEEdisplaynontitleabstractindextext

\IEEEpeerreviewmaketitle
\IEEEraisesectionheading{\section{Introduction}\label{sec:introduction}}
\IEEEPARstart{H}{uman} attentional mechanism (HAM) allows us to focus on interesting regions and filter out irrelevant ones. This cognitive ability helps us quickly understand visual scenes out of an overwhelming amount of information. Over the past decades, many works devote to imitating HAM. This task, namely saliency detection, is essential for progress in image understanding and has shown great potential in various computer vision and image processing tasks, such as image segmentation \cite{li2014the}, visual tracking \cite{hong2015online,article}, object recognition \cite{Ren2015Faster,dai2016r-fcn} and robot navigation \cite{inproceedings}.

The existing saliency detection methods can be roughly divided into three categories based on the 2D (RGB), 3D (RGB-D) and 4D (light field) input images. Different from 2D and 3D data, the light field provides multi-view images of the scene through an array of lenslets and  produces a stack of focal slices, containing abundant spatial parallax information as well as depth information. A stack of focal slices cater to human visual perception and are observed in sequence with a combination of eye movements and shifts in visual attention. Such abundant 4D data provides efficient saliency cues for saliency detection in challenging scenes such as similar foreground and background, small salient objects and complex background  \cite{li2014saliency,li2015weighted,IJCAI19}.

Light field not only brings vitality and vigor but also enriches mechanism design, and meanwhile poses challenges to saliency detection. \textbf{1)} The deficiency of light field data in terms of scale, category and element types limits the generalization of deep saliency models. \textbf{2)} Light field methods are both computation-intensive and memory-intensive as high dimensional data are employed, e.g., the model size of  the first deep-learning based light field saliency detection network is more than 119 MB and FPS is only 2 on a single 1080Ti GPU card \cite{IJCAI19}. \textbf{3)} Light field data are less ubiquitous as RGB data taken by traditional digital cameras, taking advantage of light field data in a user-friendly way for mobile devices is challenging.

Saliency detection as a pre-processing step for many tasks should be efficient, versatile and user-friendly. Building on the above observation, there are three key issues
needed to be considered: how do we construct a comprehensive dataset for achieving higher generalization accuracy of models; how do we efficiently take advantage of high dimensional light field data; how do we design the network to make it versatile enough to ensure superior performance for higher requirements on desktop computers while boosting flexibility and  productivity for mobile devices.

\begin{table*}[!t]
	\centering
	\setlength{\tabcolsep}{1mm}
	\begin{threeparttable}
		\caption{Overview of Light Field Datasets with Various Applications}
		\label{tab:dataset-overview}
		\begin{tabular}{ccccccccccc}
			\toprule
			\multicolumn{1}{c}{\multirow{2}{*}{Dataset}}&
			\multicolumn{1}{c}{\multirow{2}{*}{Year}}&
			\multicolumn{1}{c}{\multirow{2}{*}{Application}}&
			\multicolumn{1}{c}{\multirow{2}{*}{Capture}}&
			\multicolumn{1}{c}{\multirow{2}{*}{Type}}&
			\multicolumn{1}{c}{\multirow{2}{*}{Num}}&
			\multicolumn{1}{c}{\multirow{2}{*}{Class}}&
			\multicolumn{4}{c}{Image}\cr
			\cmidrule(lr){8-11}
			&{}&{}&{}&{}&{}&{}&\small RGB&\small Multiview&\small Focal Stack&\small Depth\cr
			\midrule
			HCI-Old \cite{wanner2013datasets}&\small VMV-13&\small Depth Estimation&\small Blender&\small Synthetic&12&-&$\surd$&$\surd$&$\times$&$\surd$\cr
			LFSD \cite{li2017saliency}&\small CVPR-14&\small Saliency Detection&\small Lytro I&\small Real-World&100&-&$\surd$&$\times$&$\surd$&$\surd$\cr
			LIFFAD \cite{raghavendra2015presentation}&\small TIP-14&\small Face Recognition&\small Lytro I&\small Real-World&80&-&$\surd$&$\times$&$\surd$&$\surd$\cr
			Mobile \cite{suwajanakorn2015depth}&\small CVPR-15&\small Depth Estimation&\small Samsung S3&\small Real-World&13&6&$\surd$&$\times$&$\surd$&$\surd$\cr
			TOLF \cite{Yichao2019TransCut2}&\small ICCV-15&\small Object Semantation&\small Camera Array&\small Real-World&18&7&$\surd$&$\surd$&$\times$&$\times$\cr
			HCI-New \cite{honauer2016a}&\small ACCV-16&\small Depth Estimation&\small Blender&\small Synthetic&24&-&$\surd$&$\surd$&$\times$&$\surd$\cr
			STFLytro \cite{STFLytro}&\small Stanford-16&\small Super-Resolution&\small Lytro II&\small Real-World&300&-&$\surd$&$\surd$&$\surd$&$\surd$\cr
			EPFL \cite{EPFL2016}&\small QoMEX-16&\small Super-Resolution&\small Lytro II&\small Real-World&118&10&$\surd$&$\surd$&$\times$&$\times$\cr
			LFMD \cite{LFMD2016}&\small ECCV-16&\small Material Recognition&\small Lytro II&\small Real-World&1200&12&$\surd$&$\surd$&$\times$&$\times$\cr
			HFUT \cite{zhang2017saliency}&\small TOOM-17&\small Saliency Detection&\small Lytro II&\small Real-World&255&-&$\surd$&$\surd$&$\surd$&$\surd$\cr
			LFVSD \cite{srinivasan2017learning}&\small ICCV-17&\small View Synthesis&\small Lytro II&\small Real-World&3300&7&$\surd$&$\surd$&$\times$&$\times$\cr
			LFVD \cite{sabater2017dataset}&\small CVPRW-17&\small Depth Estimation&\small Camera Array&\small Real-World&30&-&$\surd$&$\surd$&$\times$&$\surd$\cr
			INRIA \cite{INRIA2018}&\small TIP-18&\small Light Field Inpainting&\small Lytro II&\small Real-World&40&-&$\surd$&$\surd$&$\times$&$\surd$\cr
			LIFF \cite{dansereau2019liff:}&\small CVPR-19&\small Feature Detection&\small Lytro II&\small Real-World&850&30&$\surd$&$\surd$&$\times$&$\surd$\cr
			SLFD \cite{shi2019framework}&\small TIP-19&\small Depth Estimation&\small Blender&\small Synthetic&53&-&$\surd$&$\surd$&$\times$&$\surd$\cr
			DLFD \cite{shi2019framework}&\small TIP-19&\small Depth Estimation&\small Blender&\small Synthetic&43&-&$\surd$&$\surd$&$\times$&$\surd$\cr
			\multicolumn{1}{c}{\multirow{2}{*}{LF-Blur \cite{DBLPjournals}}}&\multicolumn{1}{c}{\multirow{2}{*}{\small SPL-19}}&\multicolumn{1}{c}{\multirow{2}{*}{\small Light Field Debluring}}&\small Lytro II +&\small Real-World +&\multicolumn{1}{c}{\multirow{2}{*}{400}}&\multicolumn{1}{c}{\multirow{2}{*}{-}}&\multicolumn{1}{c}{\multirow{2}{*}{$\surd$}}
			&\multicolumn{1}{c}{\multirow{2}{*}{$\surd$}}&\multicolumn{1}{c}{\multirow{2}{*}{$\surd$}}&\multicolumn{1}{c}{\multirow{2}{*}{$\times$}}\cr
			&{}&{}&\small Unreal CV&\small Synthetic&{}&{}&{}&{}&{}\cr
			DUTLF-V1 \cite{IJCAI19}&\small IJCAI-19&\small Saliency Detection&\small Lytro II&\small Real-World&1462&-&$\surd$&$\surd$&$\times$&$\times$\cr
			DUTLF-V2&\small -&\small Saliency Detection&\small Lytro II&\small Real-World&4204&102&$\surd$&$\surd$&$\surd$&$\surd$\cr
			
			\bottomrule
		\end{tabular}
	\end{threeparttable}
\end{table*}

Our work aims to shed light on these questions and make further step towards light field saliency detection. \textbf{First}, we present the largest light field dataset, extended from \cite{IJCAI19}. The DUTLF-V2 contains 4204 light field samples divided into 102 subclasses. There are multiple advantages to this extended DUTLF-V2: real-life scenarios, 1.2 times more samples than the previous largest dataset LFVSD \cite{srinivasan2017learning}, 3.4 times more classes than the previous diversest dataset LIFF \cite{dansereau2019liff:}, a variety of data elements (multiview, focal slices, depth map and all-in-focus image) being engineered to enable versatile applications of RGB, RGB-D, and light field saliency detection. 

\textbf{Second}, we propose two modules to mimic the prediction mechanism in the brain perceiving visual information, given two phrases—recruiting and screening that the eyes process all information in our visual field (Recruiting is the act of gathering visual resources. Screening is the act of using these resources to select aspects of visual information). Correspondingly,  the multi-focusness recruiting module (MFRM) is designed to recruit rich saliency features from every single focal slice for ensuring both effectiveness and diversity, while the multi-focusness screening module (MFSM) is designed to screen useful features by scanning salient objects at various locations and emphasizing the most relevant ones. This designing mechanism ensures higher-performance demand on desktop computers. 

\textbf{Third}, we propose a novel learning strategy leveraging the concept of knowledge distillation \cite{hinton2015distilling}, where triple bridges are introduced to transfer focusness knowledge from the teacher to the student. The proposed multi-focusness distillation (MFD) encourages multi-focusness consistencies between the teacher and the student; the attentive focusness distillation scheme (AFD)  mimics the attention map from the teacher; and screened focusness distillation (SFD) learns complementarity between the screened focusness knowledge from the teacher and appearance information from the student. The proposed distillation schemes ensure better absorption and integration of focusness knowledge for the student. It is noted that with the knowledge learned from the teacher, the student no longer needs the focal slices but a single RGB image as input. The lightweight and single-RGB-input student network guarantees flexibility and productivity for mobile devices.

\textbf{Furthermore}, our distillation schemes can further facilitate other existing RGB saliency detection approaches to achieve high efficiency while preserving accuracy. This makes the proposed distillation schemes a universe tool, generally suitable to a wide range of applications.

\textbf{Last but not the least},  extensive experiments on three light field datasets demonstrate the effectiveness of the proposed framework. Our teacher network achieves state-of-the-art results on three datasets, and student network (VGG16) achieves Top-2 F-measure on DUTLF-V2. The student minimizes the model size by 83\% and boosts the Frame Per Second (FPS) by 5 times, compared with the best
performing method.

An earlier version of this work was published in \cite{IJCAI19}. The specific changes implemented in this work are: \textbf{1)} We expand the DUTLF-V1 \cite{IJCAI19} to a larger-scale, higher-diversity and broader-coverage update, which can powerfully assist in
comprehensive scene understanding and higher generalization for RGB, RGB-D and light field models, and further contributes to studies in the community of saliency detection and other relative tasks.
\textbf{2)} In order to guide the student network toward an effective fusion of the multi-focusness features, we propose the attentive focusness distillation scheme that enforces the student network to mimic the attention map form the Focal stream. This distillation scheme further consistently improves the performance of our proposed method. \textbf{3)} We apply the proposed three distillation schemes to existing saliency models under two settings, including no extra parameters setting and few extra parameters setting. Extensive experiments confirm that the proposed two application settings are both easy to operate, and can be universally applied to existing saliency models while achieving significant performance gains over the original models.

\section{Related Work}

\subsection{Light Field Dataset}
Light field \cite{lumigraph,levoy1996light} contains richer visual information of the scene, and can be recorded with a camera array \cite{wilburn2005high}, a sensor equipped with lenslet array \cite{AdelsonSingle,dansereau2017a} or a coded mask \cite{marwah2013,veeraraghavan2007dappled}. It can produce ample image types that are available for various visual tasks, such as a stack of focal slices containing abundant spatial parallax information, depth maps with structural imformation and multi-view images rich in geometric information. The high-dimensional representation of light field information offers powerful capabilities for a variety of applications as shown in Table \ref{tab:dataset-overview}, including depth estimation \cite{wanner2013datasets,suwajanakorn2015depth,honauer2016a,sabater2017dataset,shi2019framework}, salient object detection \cite{li2017saliency,zhang2017saliency,IJCAI19}, face recognition \cite{raghavendra2015presentation}, super-resolution \cite{STFLytro,EPFL2016}, material recognition \cite{LFMD2016}, view synthesis \cite{srinivasan2017learning}, feature detection \cite{dansereau2019liff:}, object segmentation \cite{Yichao2019TransCut2}, light field inpainting \cite{INRIA2018} and light field debluring \cite{DBLPjournals}.

Over the past few years, several light field datasets have been introduced for salient object detection. Li \emph{et al.} \cite{li2014saliency}, \cite{li2017saliency} pioneered early work for light field dataset specially designed for salient object detection, containing 100 samples. Later on, Zhang \emph{et al.} introduced \emph{HFUT} \cite{zhang2017saliency}, a more challenging dataset with 255 real-life scenarios at various conditions. Advances in deep convolutional neural networks have enabled significant progress of a variety of large-scale saliency detection models. Although above datasets encouraged light field salient object detection to various degrees, the small size of the datasets hardly enables the deep-learning based models that generalize effectively. In addition, those datasets are constrained in terms of scene coverage and salient object diversity, failed in well generalizing in real-world applications. This left a gap in deep-learning based light field saliency detection for which inadequate information limits reviewers to verify their proposals. A larger-scale dataset \emph{DUTLF-V1} \cite{IJCAI19} remedied the above limitations to a certain extent. It contains 1462 selected high-quality samples with a variety of indoor and outdoor scenes. In our work, we take a further step to extend \emph{DUTLF-V1} to 4204 light field samples with more diverse salient objects and more various challenging scenes.

Overall, our \emph{DUTLF-V2} is clearly distinguishable from above datasets: 1) It contains a large scale of real-life scenarios captured at various distances without motion blur and uneven lighting. This can contibute to comprehensive evaluation of saliency models and pave the way for future deep-learning based light field studies. 2) Taking diverse salient objects, varied challenging scenes (e.g., similar foreground and background, complex background and multiple salient objects), random object position as well as various object size altogether makes a solid and unique foundation for real-world light field salient object detection. 3) It consists of richer image types (e.g., multiview, focal slices, depth map and all-in-focus image). This makes our dataset applicable for the field of RGB, RGB-D and light field salient object detection.
\begin{figure*}[!ht]
	\begin{center}
		\includegraphics[width=1\linewidth]{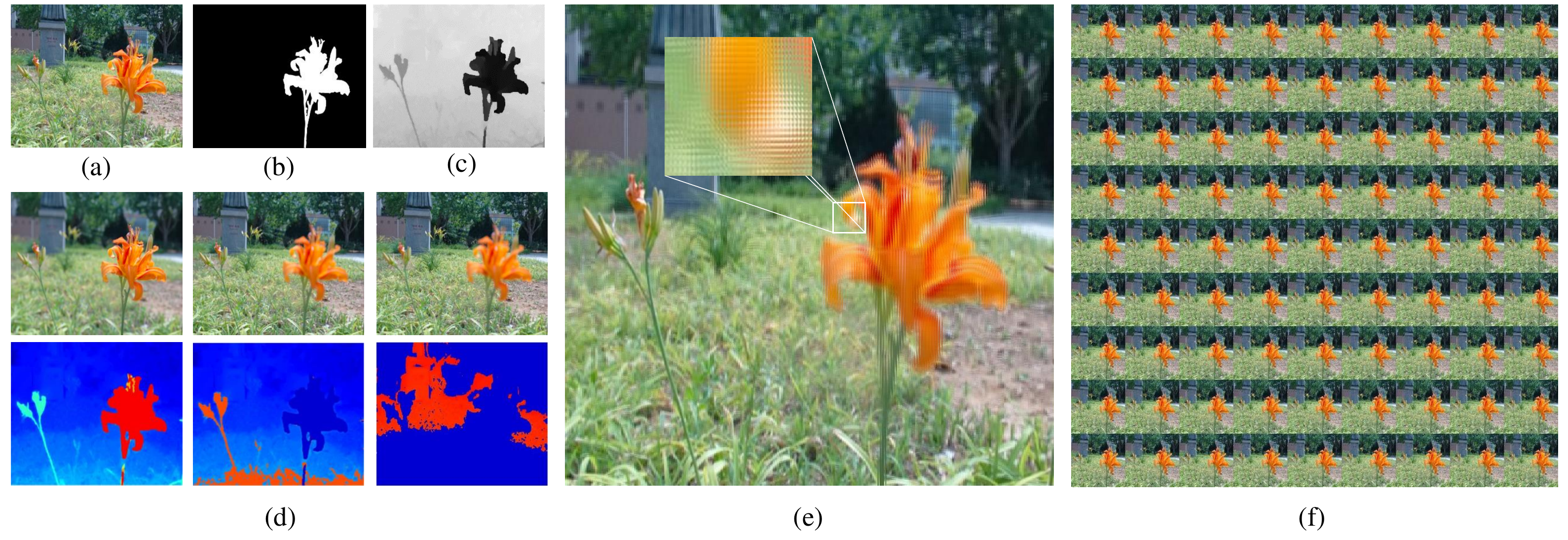}
	\end{center}
	\caption{Image elements in the proposed dataset: (a) RGB image, (b) grount truth, (c) depth map, (d) focal stack ($2^{nd}$ row) and corresponding focus region ($3^{rd}$ row), (e) micro-image and (f) multiview.}
	\label{fig:dataset-elements}
\end{figure*}
\subsection{Salient Object Detection}
Existing saliency detection models can be generally categorized as 2D (RGB), 3D (RGB-D) and 4D (light field) based approaches. Early 2D methods \cite{tu2016real,qin2015saliency,li2013saliency} focus on exploiting low-level visual cues with hand-crafted features, such as color, region contrast, etc. Benefiting from the development of convolutional neural networks, numerous 2D methods based on CNNs are proposed. Li \emph{et al.} \cite{li2018contour} developed a contour-to-saliency transferring method which can convert a pre-trained contour detection model into a saliency detection model. Liu \emph{et al.} \cite{liu2019simple} produced detail enriched saliency maps by designing two pooling-based modules which can progressively exploit high-level features. In \cite{wu2019cascaded}, Wu \emph{et al.} came up with a cascaded partial decoder which can improve both efficiency and accuracy of the existing multi-level feature aggregation networks. Chen \emph{et al.} \cite{chen2020salient} proposed a GCPANet to effectively integrate low-level appearance features, high-level semantic features and global context features. Pang \emph{et al.} \cite{MINet-CVPR2020} introduced an aggregate interaction module and a self-interaction module to generate efficient multi-level and multi-scale features. Wu \emph{et al.} \cite{wu2019scrn} proposed a stacked cross refinement network which can simultaneously refine multi-level edges and segmentation features.
In \cite{wei2020f3net}, Wei \emph{et al.} proposed a cross feature module and a cascaded feedback decoder to solve the existing big differences between features from different receptive fields.

Despite the large advance in RGB saliency detection, many scientific challenges remain for researchers to address. One of concerns is that the appearance features in RGB data are less predictive regarding challenging scenes, such as similar foreground and background, low contrast, complex background and transparent object. To address this issue by exploiting geometric and structural information, depth cues is introduced to RGB-D saliency detection. Zhu \emph{et al.} \cite{zhu2018pdnet} introduced a two-branch network in which the depth-based features generated from a sub-network are incorporated into a master network for processing RGB images. Chen \emph{et al.} \cite{chen2018progressively} proposed a complementarity-aware fusion module which can learn complementary information from the paired modalities. In \cite{chen2019multi}, Chen \emph{et al.} utilized cross-modal interactions to encourage complements across both high-level and low-level features.  Chen \emph{et al.} \cite{chen2019three} proposed a three-stream network with an attention-aware mechanism which can adaptively select complementary features.  Zhang \emph{et al.} \cite{zhang2020feature} designed a novel top-down multi-level fusion structure for effectively utilizing different fusion manners to
combine the low-level and high-level features. In \cite{chen2020rgbd}, Chen \emph{et al.} proposed a disentangled cross-modal fusion network 
for exposing structural and content representations from both modalities.

Light field data have been demonstrated in favor of saliency detection. Li \emph{et al.} \cite{li2017saliency} introduced the first light field saliency detection dataset and incorporated focusness and objectness cues into saliency detection.  Zhang \emph{et al.} \cite{zhang2015saliency} developed the background priors encoded by light field focusness to enhance the saliency and reduce the background distraction. Li \emph{et al.} \cite{li2015weighted} designed a weighted sparse coding framework which can process the heterogenous input data effectively.  Zhang \emph{et al.} \cite{zhang2017saliency} utilized multiple light field cues included all-in-focus images, depth maps, focal slices and multi-view images to exploit corresponding saliency. Piao \emph{et al.} \cite{IJCAI19} introduced a light-field-driven network to extract rich saliency representations and build the relationship between salient objects and scene understanding. Zhang \emph{et al.} \cite{Zhang_2019_NeurIPS} proposed a novel memory-oriented decoder which can comprehensively exploit internal correlation of focal slices for accurate saliency prediction. In \cite{zhang2020lfnet}, Zhang \emph{et al.} proposed a light field refinement module and a light field integration module to take full advantage of light field data.
Due to the unique property of light field, it has shown promising prospects in saliency detection.

However, higher-dimensional light field data pose new challenges in light field saliency detection, such as overcoming the memory limitation and improving computational efficiency. To confront the above difficulties, in this paper, we design an asymmetrical two-stream network in which teacher network exploits focal slices for higher-performance demand on desktop computers, while the lightweight and single-RGB-input student network aims to achieve flexibility and productivity for mobile devices.

\section{The DUTLF-V2 Dataset}
In order to provide a generic benchmark for salient object detection, we elaborately built a large-scale, high-diversity and broad-coverage dataset, namely DUTLF-V2, considered as the updated version of DUTLF-V1. The DUTLF-V2 dataset consists of 4204 light field samples, increased by 187\% from the previous version in number. It covers diverse realistic scenes corresponding to 10 main classes, which can be further divided into 102 sub-classes. Each sample of DUTLF-V2 contains 6 elements, including raw light field data, a RGB image, a saliency mask, a depth map, a stack of 12 focal slices and an array of multi-view images. A variety of data elements are engineered to enable versatile applications of DUTLF-V2 for RGB, RGB-D and light field saliency detection. The complete dataset is available at \url{https://github.com/OIPLab-DUT/DUTLF-V2}. Next, we will show details of DUTLF-V2 regarding the following 4 key aspects.
\subsection{Dataset Construction}
The construction of DUTLF-V2 includes the process of data collection, data annotation and dataset split. Implementation details are described as follows.
 
\noindent
\textbf{Data Collection.} We conducted our data collection with two main goals in mind: 1) to ensure sensory rich and logistically practical data collection, and 2) to remedy the data deficiency problem while ensuring diversity of the benchmark dataset. We opt for a commercially available Lytro Illum camera for our data collection, being consistent with previous works \cite{li2014saliency}, \cite{IJCAI19}, \cite{li2017saliency}, \cite{zhang2017saliency}. Our dataset consists of a variety of indoor and outdoor scenes captured in the surrounding environments, e.g., offices, supermarkets, campuses, streets and so on. Moreover, our dataset is recorded over several months under different time, lighting conditions and camera parameters (e.g., aperture size and focal length that determine depth of field in an image). To assure the quality of the data, three participants were employed full time to screen out disqualified images, such as blurred images and images with disputable salient objects. In this way, 4204 light field samples are included in DUTLF-V2. 

\noindent
\textbf{Data Annotation.} To ensure the accuracy and consistency for annotation, twenty participants are instructed to label the salient objects from the all-in-focus RGB image using a widely-used custom segmentation tool. To further improve annotation accuracy, all participants are pre-trained with over ten examples. To achieve a consensus, each three of participants jointly determined the saliency objects, and cross validated by other three ones. We accept a consensus of at least 83 percent to be considered a positive label.
In the annotation process, the participants were first asked to draw the coarse boundary along the salient objects, then check the segmentation results and refine the boundaries.
To this end, we obtained corresponding 4204 accurate pixel-wise ground-truth masks in total.

\noindent
\textbf{Dataset Split.} To ensure evaluation consistency, the split protocol of our dataset is similar with previous works \cite{dataset-duts}, \cite{dataset-dutrgbd}. 4204 light field samples are randomly generated into the training and testing sets at the ratio of 7:3. Specifically,  2957 samples are for training  and 1247 samples are for testing with corresponding saliency masks. The proportion of hard samples in the training set and test set is 57.25\% and 60.22\%, respectively.  We note that simple and hard samples are well balanced to split into the training and testing sets.

\begin{figure*}[!ht]
\begin{center}
  \includegraphics[width=1\linewidth]{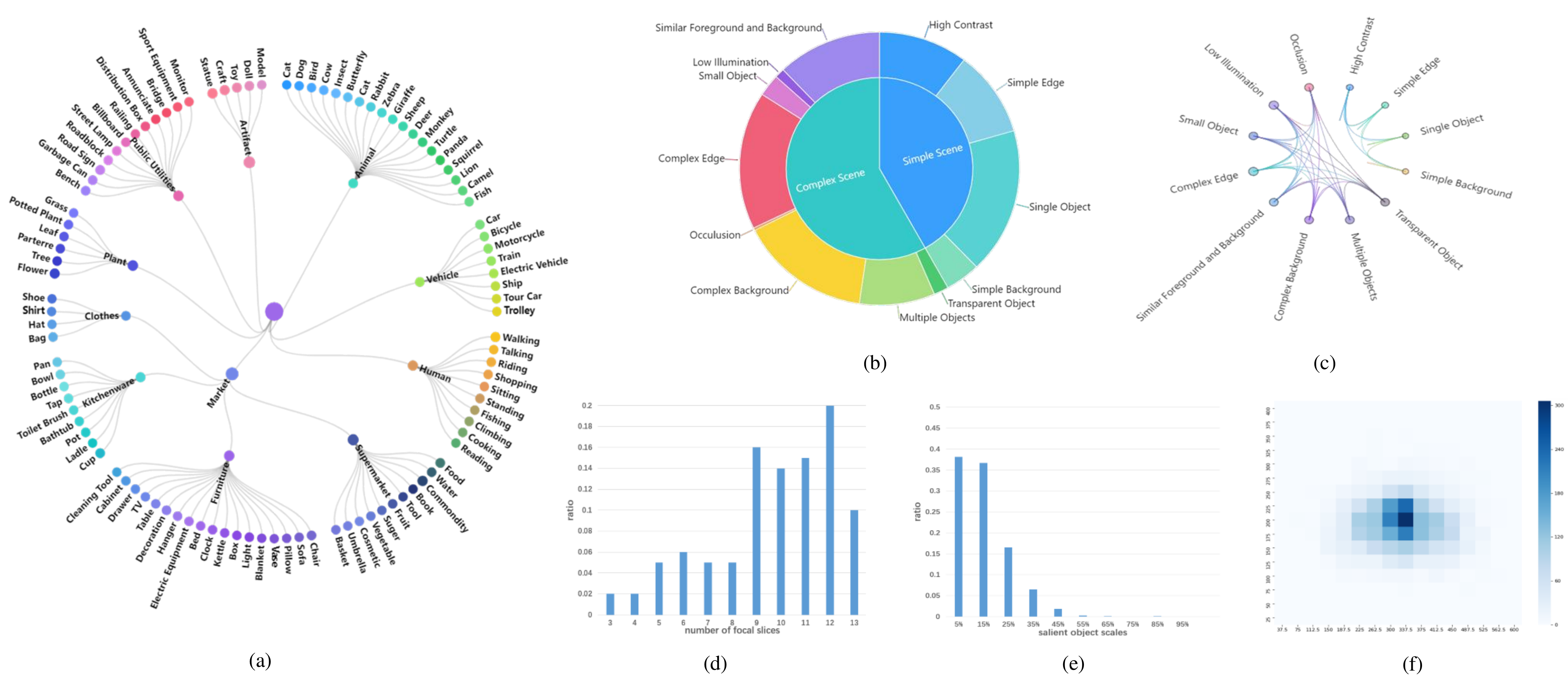}
\end{center}
   \caption{Dataset statistics: (a) Salient objects categories, including 10 main classes and 102 sub-classes. (b) Scene categories, including complex scenes and simple scenes, accounting for 58.14\% and 41.86\%, respectively. (c) Mutual dependencies among scene categories in (b). (d) Ratio distribution according to number of slices in each focal stack, the coordinate axis x and y represent the number of focal slices and coresponding ratio, respectively. (e) Ratio distribution according to salient object scales, the coordinate axis x and y represent the salient object scales and coresponding ratio, respectively. (f) Center distribution of the salient objects depicted in an image coordinate system.}
\label{fig:dataset-features}
\end{figure*}

\subsection{Dataset Elements}
To achieve versatility of our dataset, we decode the light field raw data using the Lytro Power Tools (LPT). Each light field consists of a RGB (all-in-focus) image, a corresponding manually labeled ground truth, a depth image, a stack of focal slices focusing at different depths and an array of multi-view images. We will elaborate on each element of DUTLF-V2, illustrated in Figure \ref{fig:dataset-elements}. 

\noindent
\textbf{Raw Light Field Data.} Light field data are stored in raw files, which save the data that come off the image sensor, namely Light Field Raw (LFR) files. The raw light field data comprise an array of micro-images shown in Figure \ref{fig:dataset-elements} (e), each micro image being projected by a respective micro lens of a micro-lens array. LFR files can be used as initial input for either the Lytro camera software or any other processing toolbox. In our implementation, we use the Lytro Power Tools (LPT) to decode the LFR files into RGB (all-in-focus) images, depth maps, focal stacks and multi views. 

\noindent
\textbf{RGB Image (All-in-Focus).} An all-in-focus image has all parts of the sample in focus. It can be generated using a Markov random field (MRF) based approach by integrating the sharpest in-focus pixels across the focal stack. We obtained 4204 all-in-focus images with the resolution of $600\times400$, which are applicable for RGB saliency detection.

\noindent
\textbf{Saliency Mask.} To obtain the pixel-wise ground truth, we manually labeled the salient objects from the RGB image using a commonly used segmentation tool. Three annotators are required to determine the salient objects in order to achieve annotation consensus.

\noindent
\textbf{Depth Map.} Each pixel in the depth map describes the distance from a viewpoint to the surfaces of scene objects. Depth maps can be generated using the focal stack. To highlight the relative position between objects, we normalize the depth map to the range from 0 to 255. The depth maps together with RGB images, enable versatile application of our dataset for RGB-D saliency detection.

\noindent
\textbf{Focal Stack.} A focal stack represents a series of images focusing at different depths. It caters to human visual perception and is observed in sequence with a combination of eye movements and shifts in visual attention. In our proposed dataset, the focal stack has different number varied from 3 to 13 (see Figure \ref{fig:dataset-features}(d)). Most of the focal stacks contain more than 9 slices, which can demonstrate the diversity of image depths.
We ensure the number of focal slices in each scene to be 12 by randomly copying for coding requirements. This allows our dataset to be effectively applied in light field salieny detection.

\noindent
\textbf{Multi View / Micro Image.} The light field provides multi-view images of the scene through an array of lenslets, namely multiview. The multiview is a set of sub-view images, and each 2D sub-view image 'sees' the scene from a slightly different viewpoint, thus providing angular variation which is the distinctive characteristic of light field imaging. In our DUTLF-V2, each multiview contains 81 sub-view images, which have a 9$\times$9 angular resolution and 600$\times$400 pixels of spatial resolution. The sufficient multiview can facilitate the research of light field saliency detection, super-resolution and view synthesis.
\subsection{Dataset Statistics}
Our motivation for DUTLF-V2 is to build a versatile dataset in large scale, high diversity and broad coverage. We conduct descriptive statistics on our DUTLF-V2 in terms of the following aspects.

\noindent
\textbf{Diversity of Salient Objects.} High-quality datasets should reflect the diversity in a broad range of coverage. Our DUTLF-V2 covers a variety of indoor and outdoor scenes that enable a comprehensive understanding of the visual systems. 
Our dataset can be categorized into 10 main classes such as animal, vehicle, plant, public utilities, artifact, clothes, kitchenware, furniture, supermarket and human, and 102 sub-classes, such as animal (e.g., cat, insect, sheep), vehicle (e.g., bicycle, car), plant (e.g., tree, flower), public utilities (e.g., bench, road sign), etc., shown in Figure \ref{fig:dataset-features}(a).
Besides, our dataset reflects the daily situation in a realistic way, further taking practical application into consideration.
\begin{figure*}[t]
	\begin{center}
		\includegraphics[width=1\linewidth]{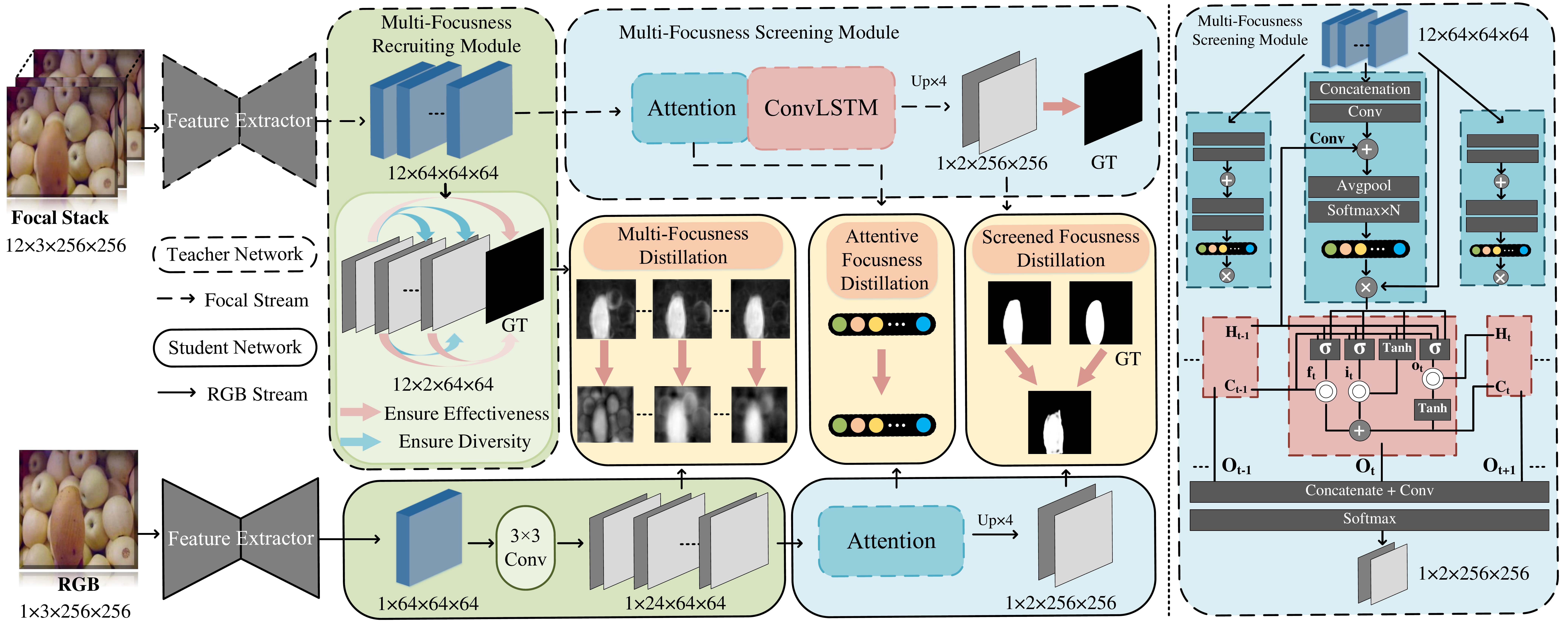}
	\end{center}
	\caption{The whole pipeline. Our asymmetrical two-stream network consists of the Focal stream and the RGB stream. The Focal stream, served as the teacher network, contains a feature extractor, a Multi-Focusness Recruiting Module(MFRM) and a Multi-Focusness Screening Module(MFSM). The lightweight student network learns to replace focal slices with a single RGB image via three tailored distillation schemes.}
	\label{fig:pipeline}
\end{figure*}

\noindent
\textbf{Challenging Scenes.} A certain amount of hard samples in datasets may guarantee generalization ability of deep models. With this in mind, we aim to build a more challenging and representative dataset. Figure \ref{fig:dataset-features}(b) represents a variety of challenging scenes in our DUTLF-V2, such as similar foreground and background, low illumination, small object, complex edge, occlusion, complex background, multiple objects and transparent object. 
It can also be seen that the complex scenes have a larger share in our dataset than the simple scenes. 
More importantly, there are mutual dependencies among scene categories shown in Figure \ref{fig:dataset-features}(c). That means a scene may belong to multiple complex categories (e.g., a scene consists of multiple salient objects under low illumination).
The above analysis demonstrates that the proposed DUTLF-V2 is challenging and presents more challenges to the field of salient object detection.

\noindent
\textbf{Scale of Salient Objects.} The size of salient objects refers to the proportion of foreground object pixels over the whole image. Different scale of salient objects is another factor to determine if the datasets are high qualified. Figure \ref{fig:dataset-features}(e) represents that the ratio distribution according to salient object scales lies in a broad range of [0.001, 0.865] and most objects occupy less than 35$\%$ area of the image. This attribute further contributes to higher diversity of our dataset.

\noindent
\textbf{Distribution of Center bias.} Location of salient objects presents another challenge in that datasets simply composing centered salient objects may not be sufficient to generically evaluate saliency models.
Figure \ref{fig:dataset-features}(f) shows the location distribution for the center of salient objects across the whole dataset. It can be seen that the salient objects in our DUTLF-V2 appear at a variety of locations, which avoids the salient objects being easily extracted by the center.

\section{Method}

\begin{figure}[!ht]
	\begin{center}
		\includegraphics[width=8.5cm]{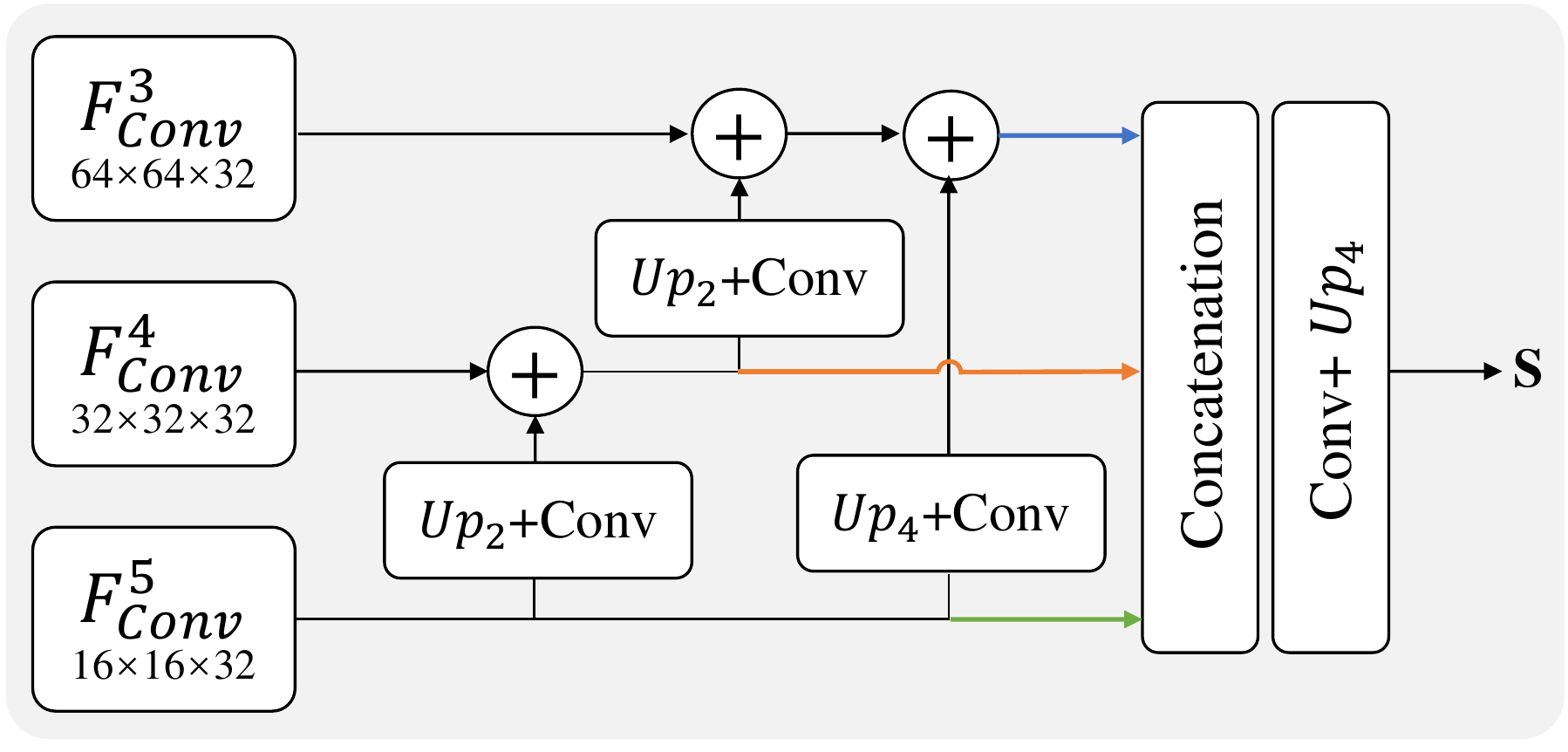}
	\end{center}
	\caption{Detailed structure of the feature extractor in the Focal
		stream or RGB stream.}
	\label{fig:decoder}
\end{figure}

To achieve the goal of developing an efficient, versatile and user-friendly architecture for light field saliency detection, we introduce an asymmetrical two-stream network based on knowledge distillation, as illustrated in Figure \ref{fig:pipeline}. On the one hand, the Focal stream, served as the teacher network, aims to learn to exploit focal slices for higher-performance demand on desktop computers. On the other hand, the student network learns to replace focal slices with a single RGB image for flexibility and productivity on mobile devices. The feature extractor in the teacher network is based on VGG19 \cite{simonyan2014very}, while the student feature extractor is based on VGG16 \cite{simonyan2014very}. We select the high-level convolutional features ($F^3_{Conv}$, $F^4_{Conv}$ and $F^5_{Conv}$) to detect salient objects. The detailed structure of the feature extractor is shown in Figure \ref{fig:decoder}. In this section, we will discuss the reasoning Focal stream (Learning to Exploit Focal Stack), and RGB stream (Learning to Replace Focal Stack) in detail.

\subsection{Learning to Exploit Focal Stack}
In order to enable our teacher network with more accurate prediction for higher-performance demand on desktop computers, as well as efficient transfer of rich focusness knowledge to the student, we propose two tailored modules in the teacher network, which are multi-focusness recruiting module (MFRM) and multi-focusness screening module (MFSM). The MFRM focuses on explicitly recruiting saliency information from each focal slice, and the MFSM aims to select useful features and suppress the unnecessary ones. Detailed discussion about the effcet of MFRM and MFSM are provided in ablation studies.
Next, we will elaborate each component in the teacher network.

\subsubsection{Multi-Focusness Recruiting Module (MFRM).}
Inspired by the recruiting phase in the brain perceiving visual information, we aim to gather rich saliency features by processing every single focal slice. A straightforword solution to this issue is supervising the raw features with ground truth for avoiding the distraction of non-salient objects and ensuring the effectiveness of each focusness feature. However, this strategy could reduce diversity and complementarity between multi-focusness features. To this end, we design a multi-focusness recruiting module (MFRM) which encourages the raw multi-focusness features containing adequate effectiveness and diversity to achieve optimal results. The detail of MFRM is shown in Figure \ref{fig:pipeline}. We first connect a convolutional layer to convert each focusness feature from 64 channels to 2 channels. Then each focusness feature is supervised by the following loss function:
\begin{equation}
\begin{array}{l}
L_R({f_k }) = L_{CE} (f_k ,Y) - \lambda \sum\limits_{i = 1,i \ne k}^N {L_{MSE} (f_k ,f_i)},
 \end{array}
\end{equation}
where $f_k$ represents the $k^{th}$ focusness feature, $Y$ represents the ground truth and $N$ represents the total number of focal slices. $L_{CE}$ and $L_{MSE}$ denote cross-entropy and mean squared error loss function, respectively. The first item encourages effectiveness, the second item enhances diversity, and the non-negative weight $\lambda$, which is set to 10, expresses the trade-off between these two items.

\subsubsection{Multi-Focusness Screening Module (MFSM).}


Inspired by the screening phase in the brain perceiving visual information, which concerns with selectivity, we aim to efficiently select useful saliency information from multi-focusness features. Therefore, by assigning different weights to different focal slices regarding the salient objects, we propose a multi-focusness screening module (MFSM) to resemble the screening phase how human select information of interest from visual resources. Specifically, the MFSM consists of a ConvLSTM model with an attention mechanism as shown in Figure \ref{fig:pipeline}. The attention module is designed to emphasize the useful features and suppress the unnecessary ones in order to produce screened features. The ConvLSTM module aims to summarize the spatial information from the screened features of historical steps and the current step for accurately identifying the salient objects. As time step increases, the MFSM can highlight the salient regions and block the non-salient ones gradually (see Figure \ref{fig:MFSM}). Detailed operations are described as follows.

In each time step $t$, the multi-focusness features $f=\{f_1, f_2, \cdots, f_N \}$ first go through a feature-wise attention module and this procedure can be defined as:
\begin{equation}
\resizebox{.99\linewidth}{!}{$\tilde F_t\!=\!\sum\limits_{i = 1}^N {f  \odot } (\Phi (AvgPool(Cat\left[ {f_1 ;f_2 ; \cdots ;f_N }\right] * W_f  + H_{t - 1}  * W_h )))$},
\end{equation}
where $\Phi$ represents softmax function. $\odot$ and $*$ mean feature-wise multiplication and convolution operation, respectively. $H_{t - 1}$ represents the hidden state of the ConvLSTM cell in the $(t - 1)^{th}$ step. $W_f$ and $W_h$ are parameters of the convolutional kernels. Then the combined features $\tilde F_t$ are fed into a ConvLSTM cell. The internal operations in ConvLSTM are shown in Figure \ref{fig:pipeline}. After several time steps, we concatenate the $O$ gates to summarize the saliency information from the screened features, and make a final prediction. This procedure can be defined as:
\begin{equation}
S_{tea} {\rm{ = }}\Phi \left( {W_s  * Cat[O_1 ; \cdots ,O_t ; \cdots ;O_T ]} \right),
\end{equation}
where $S_{tea}$ represents the saliency map predicted from the teacher network, $T$ denotes the total number of time steps and is set to 4. $W_s$ denotes the convolution parameter.

\subsection{Learning to Replace Focal Stack}
Most existing methods based on knowledge distillation take as same input for teacher and student networks. Considering simplifying the heavy focal computation and  accessing focal slices more conveniently, we aim to design a lightweight network in a user-friendly way by taking a single ubiquitous RGB image as input to replace focal slices. However, directly transferring the output from the teacher to the student overlooks the inherent differences between two types of data. Therefore, we propose three tailored distillation schemes to replace focal slices with a single RGB image by transferring the focusness knowledge. In this way, the focusness knowledge is defined as three parts: 1) The first part is designed to mimic multi-focusness features only using a single RGB image. 2) The second part is designed to learn to effectively integrate the multi-focusness features. 3) The third part is designed to learn complementary information from appearance and screened focusness knowledge. We also give an in-depth discussion about the effect of the proposed three distillation schemes in ablation studies. Detailed description for each distillation scheme is given below.

\subsubsection{Multi-Focusness Distillation Scheme (MFD).}
Unlike directly enforcing the student to mimic the output from the teacher, the student is first trained to hallucinate multi-focusness features from the Focal stream by our proposed multi-focusness distillation scheme. This is mainly driven by the consideration of the inherent differences between input data for the teacher and student. Moreover, multi-focusness features can be produced from a single RGB input without explicit focal computation, whis leads to significantly faster inference. In detail, we reduce the Kullback-Leibler divergence loss between features of the penultimate layer in RGB stream and features generated from the MFRM in Focal stream:
\begin{equation}
L_{MFD} {\rm{ = }}\frac{{\rm{1}}}{N}\sum\limits_{i = 1}^N {KL\left( {f_i^s \left\| {f_i^t } \right.} \right)},
\end{equation}
where $f_i^t$ represents the feature map of the $i^{th}$ focal slice produced from the teacher network and $f_i^s$ represents the $i^{th}$ feature map produced from the student network.
\subsubsection{Attentive Focusness Distillation Scheme (AFD).}
After learning multi-focusness features from the Focal stream, a crisis problem for the student network is how to effectively integrate the multi-focusness features. Based on above consideration, we propose the attentive focusness distillation scheme by which the student is capable to mimic the attention map from the Focal stream. In this way, the useful multi-focusness features are highlighted while the unnecessary ones are suppressed, further leading to more accurate saliency detection results. Specifically, we train the student network by minimizing the Kullback-Leibler divergence loss between the attention map of the RGB stream and the MFSM in Focal stream. This loss can be described as the following equation:
\begin{equation}
L_{AFD} {\rm{ = }}{KL\left( {A_{stu} \left\| {A_{tea} } \right.} \right)},
\end{equation}
where $A_{tea}$ denotes the attention map of the RGB stream, and $A_{stu}$ denotes the attention map of the MFSM in Focal stream.
\subsubsection{Screened Focusness Distillation Scheme (SFD).}
Our third distillation scheme goes a further step: we align the class probability of each pixel produced from the student network and the teacher network, as well as the probability of each pixel between the output of the student network and the ground truth. We refer this distillation scheme as screened focusness distillation, which allows the student network to learn complementary information from appearance and screened focusness information for accurate prediction. To enhance this process with screened focusness and appearance information, we train the student network by backpropagating a linear combination of KL and cross entropy losses through the entire network. The loss function is given as follows:
\begin{equation}
L_{SFD} {\rm{ = }}{KL\left( {S_{stu} \left\| {S_{tea} } \right.} \right)}+ \alpha L_{CE} (S_{stu} ,Y),
\end{equation}
where $S_{tea}$ and $S_{stu}$ represent the saliency map predicted from the teacher and student networks, respectively. The hyperparameter $\alpha$ is set to 1.

\subsection{Training Process}
As presented in Algorithm 1, the teacher network is supervised by two losses: the cross entropy loss $L_{CE}$ with the ground truth and the recruiting loss $L_R{(f_k)}$ in Eq.(1). During the knowledge distillation process, the teacher is pre-trained and the parameters are kept frozen. The student is supervised by a combination of the multi-focusness distillation loss $L_{MFD}$ in Eq.(4), the attentive focusness distillation loss $L_{AFD}$ in Eq.(5) and the screened focusness distillation loss $L_{SFD}$ in Eq.(6). $W_T$ and $W_S$ are parameters for the teacher and student, respectively.

\begin{algorithm}[!ht]
\caption{Training Process of Our Proposed Method}
\textbf{Stage 1 :} {Training the teacher network.}

\textbf{Inputs :} {Focal slices.}

\hspace{1em}$W_T {\rm{ = argmin}}_{W_T } \left( {L_{CE} \left( {S_{tea}, Y } \right) + \sum\limits_{k = 1}^N {L_R \left( {f_k } \right)} } \right)$

\textbf{Stage 2 :} {Training the student network.}

\textbf{Inputs :} {Single RGB.}

\hspace{1em}$W_S {\rm{ = argmin}}_{W_S } \left( {L_{MFD}  + L_{AFD} + L_{SFD} } \right)$
\end{algorithm}

\section{Experiments and Analyses}
In this section, we first presented the adpoted datasets, evaluation metrics and implementation details. Then we gave a comprehensive comparision of our DUTLF-V2 with other RGB, RGB-D and light field benchmarks. Next we performed a number of ablation experiments to analyze the effcet of each component in teacher and student networks. We also comprehensively compared the performance of our approach with other advanced saliency models. Besides, detailed studies are provided about the application of our proposed distillation schemes under two settings.
\subsection{Experimental Setup}

\subsubsection{Benchmark Datasets.}
We evaluated our approach on the proposed DUTLF-V2 and other two public light field datasets: HFUT-LFSD \cite{zhang2015saliency} and LFSD \cite{li2014saliency}. DUTLF-V2 consists of 4204 light field samples under a wide range of indoor and outdoor scenes which can be divided into 102 classes. HFUT-LFSD and LFSD include 255 and 100 light fields, respectively. Each light field consists of an all-in-focus image, 12 focal slices focused at different depths and a corresponding manually labeled ground truth.

In order to achieve a fair comparison, we randomly genearted 2957 samples from DUTLF-V2 dataset and additional 100 samples from HFUT-LFSD dataset as the training set. The remaining samples of DUTLF-V2 and HFUT-LFSD, and the LFSD dataset are used for testing. To avoid overfitting, we augmented the training set by flipping, cropping and rotating.

\subsubsection{Evaluation Metrics.}
To evaluate the performance of our approach and other methods, we adopted 5 widely-used metrics:  F-measure \cite{achanta2009frequency}, weighted F-measure \cite{margolin2014evaluate}, Mean Absolute Error (MAE), S-measure \cite{fan2017structure}, and E-measure \cite{fan2018enhanced}. The above evaluation metrics can provide comprehensive and reliable evaluation results.
Specifically, the F-measure can comprehensively evaluate the quality of saliency maps and the weighted F-measure is adopted to overcome the interpolation flaw, dependency flaw and equal-importance flaw for a fair comparison. The MAE computes the average absolute per-pixel difference between the saliency map and corresponding ground truth. The S-measure can evaluate the spatial structure similarities and the E-measure can jointly capture image level statistics and local pixel matching information. In addition, we adopted model size and Frames Per Second (FPS) to evaluate the complexity of each method.

\subsubsection{Implementation Details.}

Our method is implemented with Pytorch toolbox and trained on a PC with GTX 1080Ti GPU. We train both teacher network and student network using the SGD optimization algorithm in which the momentum, weight decay and learning rate are set to 0.9, 0.0005 and 1e-10, respectively. The hyperparameter temperature T is set to 20 in all distillation loss functions. The minibatch size is 1 and maximum iterations of both the teacher network and student network are set to 500000.

\subsection{Comparison of Benchmarks}

In order to offer deeper insights into the proposed dataset, we compared the DUTLF-V2 to other RGB, RGB-D and light field datasets.

\subsubsection{Comparison with RGB Datasets}
Most existing RGB datasets include SOD \cite{movahedi2010design}, ECSSD \cite{dataset-ecssd}, PASCAL-S \cite{dataset-pascal}, DUTS \cite{dataset-duts} and MSRA10K \cite{dataset-msra10k}. Although a considerable number of samples are contained in those datasets, they are either randomly selected from the public databases or collected from the Internet. 
In contrast, the samples of our DUTLF-V2 are collected in our daily lives. Our dataset is more closely relevant to real-life scenarios. This offers an advantage in practical application for the task of saliency detection.

\subsubsection{Comparison with RGB-D Datasets}
We compared our DUTLF-V2 with existing RGB-D datasets, including NJUD \cite{dataset-njud}, NLPR \cite{dataset-nlpr}, RGBD135 \cite{dataset-rgbd135}, DUT-RGBD \cite{dataset-dutrgbd}, SSD \cite{dataset-ssd} and STEREO \cite{dataset-stereo}.
It is noted that our DUTLF-V2 shows an advantage in scale over all the above datasets, numerically increased by 111\% compared to the current largest RGB-D dataset NJUD. This significant increase in scale is beneficial for evaluating deep saliency models.
Additionally, compared to some synthetic or indoor datasets (e.g., SSD and RGBD135), the DUTLF-V2 shows strength in coverage and generalization. Specifically, SSD contains 80 samples collected from stereo movies and RGBD135 contains 135 samples collected from constrained indoor scenes.
While our DUTLF-V2 consists of 4204 samples collected from a variety of indoor and outdoor scenes that can be divided into 102 classes. 
The introduce of our DUTLF-V2 can contribute to facilitate the research and practices of RGB-D saliency detection.

\subsubsection{Comparison with Light Field Datasets for Various Tasks}
Compared with the public light field datasets for various tasks listed in Table \ref{tab:dataset-overview}, our proposed DUTLF-V2 is significantly improved in terms of scale, diversity and applicability. We can see from Table \ref{tab:dataset-overview} that the scale of our DUTLF-V2 exceeds all the public light field datasets, increased by 27\% compared to the current largest light field dataset LFVSD \cite{srinivasan2017learning} for view synthesis. We also compared the coverage which helps to showcase the high diversity of our dataset. Our DUTLF-V2 contains 102 classes of samples, over three times that of the most diverse one \cite{dansereau2019liff:} which contains 30 classes. Additionally, we can note that DUTLF-V2 provides a variety of data elements (e.g., RGB images, multi views, focal stacks, depth maps), while over 88 percent of the listed datasets contain only three or less data types. The sufficient representation of light fields enables versatile application of our dataset in RGB, RGB-D and light field saliency detection.

\subsubsection{Comparison with the Previous Version (DUTLF-V1)}
DUTLF-V1 \cite{IJCAI19} is the existing largest light field saliency detection dataset and encouraged light field saliency detection to a certain extent. Compared with DUTLF-V1, our newly proposed DUTLF-V2 has been greatly improved in several aspects, including scale, diversity and complexity. Specifically, DUTLF-V2 contains 4204 light field samples, increased by 187\% compared to DUTLF-V1 as shown in Table \ref{tab:dataset-overview}. Meanwhile, our DUTLF-V2 covers a wider range of indoor and outdoor scenes, which can be divided into 102 classes. It is also noted that the number and type of challenging scenes are greatly increased in DUTLF-V2, meeting comprehensive training needs as well as bringing up more challenges for salient object detection. 
The numerical results of our method trained on DUTLF-V1 and  DUTLF-V2 also confirm that our extended DUTLF-V2 can powerfully assist in higher generalization of saliency models to other datasets.
As shown in Table \ref{tab:v1-v2}, the proposed method trained on DUTLF-V2 achieves considerable performance gains compared to that trained on DUTLF-V1. Specifically, in terms of the student network, the MAE is reduced by 24.1\% and 9.9\% on HFUT and LFSD, respectively. We can see that DUTLF-V2 shows advantages in training saliency models and encourages more advanced research in this field.
\begin{table}[t]
	\centering
	\setlength{\tabcolsep}{0.7mm}
	\begin{threeparttable}
		\caption{Comparision of the Training Set of DUTLF-V1 and DUTLF-V2 (\textbf{\textcolor{red}{red bold:}} top-1 results) 
		}
		\label{tab:v1-v2}
		\begin{tabular}{p{1cm}<{\centering}p{1.4cm}<{\centering}p{0.75cm}<{\centering}p{0.75cm}<{\centering}p{0.75cm}<{\centering}p{0.75cm}<{\centering}}
			\toprule
			\multicolumn{1}{c}{\multirow{2}{*}{methods}}&
			\multicolumn{1}{c}{\multirow{2}{*}{Training Set}}&
			\multicolumn{2}{c}{HFUT}&\multicolumn{2}{c}{LFSD}\cr
			\cmidrule(lr){3-4} \cmidrule(lr){5-6}
			&{}&$F_{\beta}\uparrow$&MAE$\downarrow$&$F_{\beta}\uparrow$&MAE$\downarrow$\cr
			\midrule
			\multirow{2}{*}{Teacher}&DUTLF-V1&.705&.082&.842&\textcolor{red}{\bf.080}\cr
			&DUTLF-V2&\textcolor{red}{\bf.753}&\textcolor{red}{\bf.069}&\textcolor{red}{\bf.856}&.084\cr
			\midrule
			\multirow{2}{*}{Student}&DUTLF-V1&.668&.091&.712&.151\cr
			&DUTLF-V2&\textcolor{red}{\bf.714}&\textcolor{red}{\bf.069}&\textcolor{red}{\bf.738}&\textcolor{red}{\bf.136}\cr
			\bottomrule
			\end{tabular}
			\end{threeparttable}
			\end{table}
\begin{table}[t]
	\centering
	\setlength{\tabcolsep}{0.7mm}
	\begin{threeparttable}
		\caption{Quantitative Results of the Ablation Analysis for Our Teacher Network (\textbf{\textcolor{red}{red bold:}} top-1 results) }
		\label{tab:teacher-ablation}
		\begin{tabular}{ccp{1cm}<{\centering}p{1cm}<{\centering}p{1cm}<{\centering}p{1cm}<{\centering}}
			\toprule
			\multicolumn{1}{c}{\multirow{2}{*}{Module}}&
			\multicolumn{1}{c}{\multirow{2}{*}{Model}}&
			\multicolumn{2}{c}{DUTLF-V2}&\multicolumn{2}{c}{HFUT-LFSD}\cr
			\cmidrule(lr){3-4} \cmidrule(lr){5-6}
			&{}&$F^{w}_{\beta}\uparrow$&MAE$\downarrow$&$F^{w}_{\beta}\uparrow$&MAE$\downarrow$\cr
			\midrule
			\multirow{2}{*}{MFRM}&Only $L_{CE}$&.769&.054&.664&.073\cr
			&w/MFRM&\textcolor{red}{\bf.792}&\textcolor{red}{\bf.050}&\textcolor{red}{\bf.687}&\textcolor{red}{\bf.069}\cr
			\midrule
			\multirow{3}{*}{MFSM}&Step 1&.740&.059&.638&.076\cr
			&Step 2&.746&.058&.653&.075\cr
			&Step 3&.778&.053&.672&.071\cr
			&Ours (step 4)&\textcolor{red}{\bf.792}&\textcolor{red}{\bf.050}&\textcolor{red}{\bf.687}&\textcolor{red}{\bf.069}\cr
			\bottomrule
		\end{tabular}
	\end{threeparttable}
\end{table}

\begin{figure}[!ht]
	\begin{center}
		\includegraphics[width=8cm]{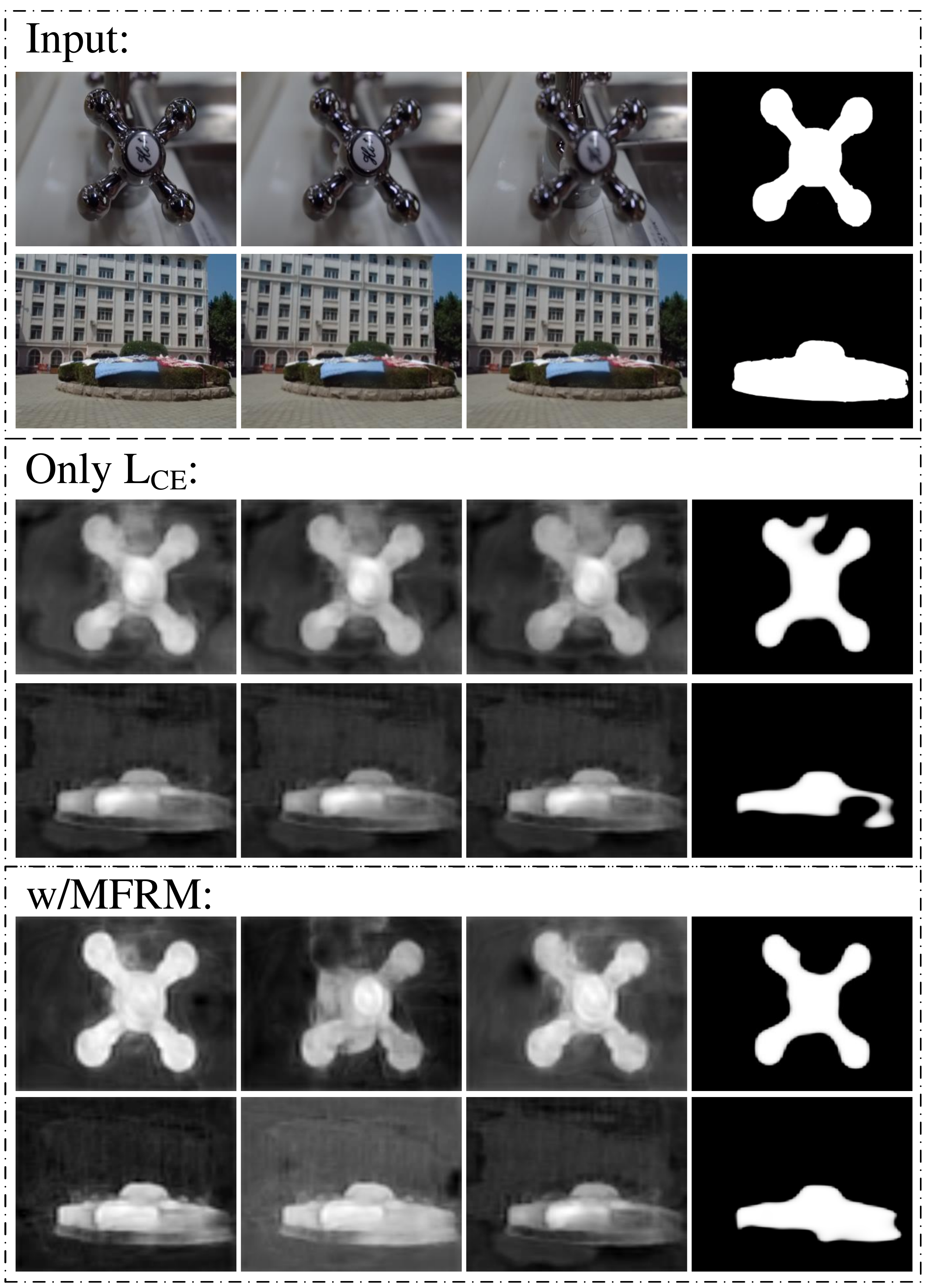}
	\end{center}
	\caption{Visual comparisions in MFRM. The 1$^{st}$ to 3$^{rd}$ columns show the focal slices and corresponding multi-focusness features. The 4$^{th}$ column shows the ground truth and saliency maps.}
	\label{fig:MFRM}
\end{figure}
\begin{figure}[!ht]
	\begin{center}
		\includegraphics[width=8.5cm]{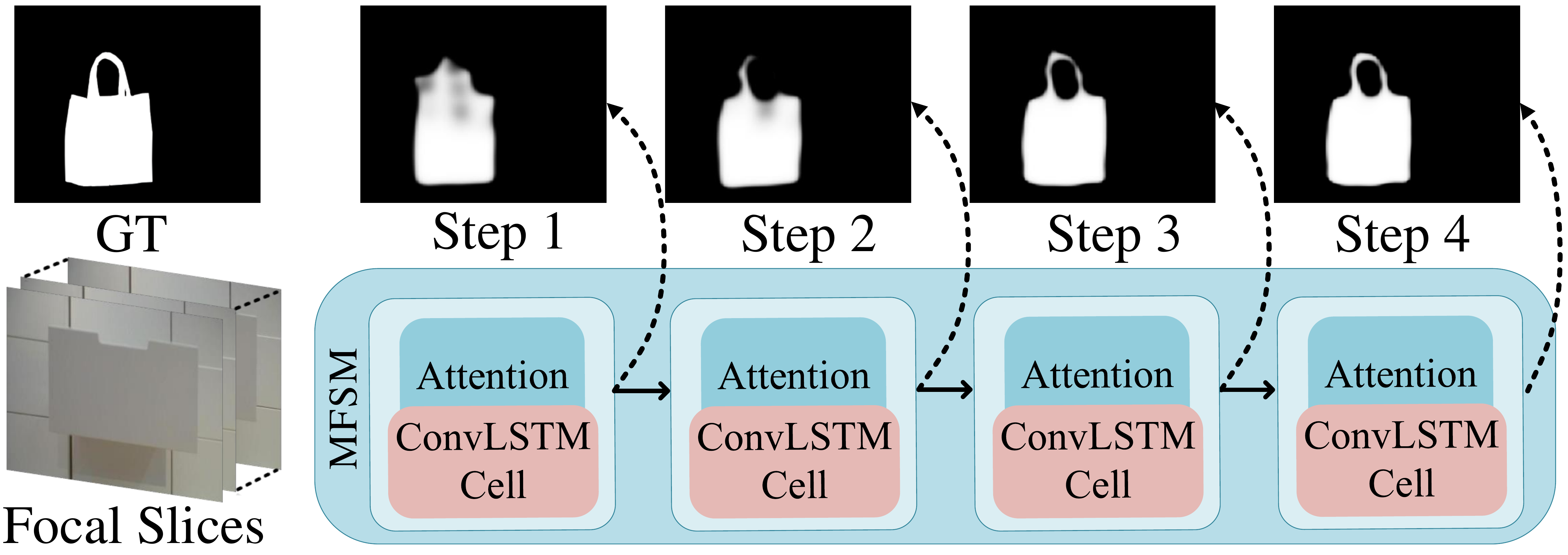}
	\end{center}
	\caption{Visual comparisions of different steps in MFSM.}
	\label{fig:MFSM}
\end{figure}
\subsection{Teacher Network Analysis}
\subsubsection{Effect of MFRM.}
To validate the MFRM of the teacher network in recruiting comprehensive saliency information from focal slices, we conducted visual comparisons (see Figure \ref{fig:MFRM}) of the multi-focusness features generated with simple supervision (denoted as `Only $L_{CE}$') and our proposed MFRM (denoted as `w/MFRM').
We can observe that the multi-focusness features generated by simple supervision are very similar. This may limit the diversity of features and lead to sub-optimal results, such as false positives (row 3) or incomplete detection of salient objects (row 4).
In contrast, the MFRM encourages adequate diversity between features of different focal slices to achieve optimal results (row 5 and row 6).
Quantitative results shown in Table \ref{tab:teacher-ablation} are numerically consistent that the MFRM brings considerable performance gains. Our MFRM reduces the MAE by 7.4\% and 5.5\% on DUTLF-V2 dataset and HFUT-LFSD dataset, respectively.
\begin{table}[t]
	\centering
	\setlength{\tabcolsep}{0.3mm}
	\begin{threeparttable}
		\caption{Ablation Analysis of the Proposed Distillation Schemes on Student Network (\textbf{\textcolor{red}{red bold:}} top-1 results) }
		\label{tab:ablation-student}
		\begin{tabular}{p{2.3cm}<{\centering}p{1cm}<{\centering}p{1cm}<{\centering}p{1cm}<{\centering}p{1cm}<{\centering}p{1cm}<{\centering}p{1cm}<{\centering}}
			\toprule
			\multicolumn{1}{c}{\multirow{2}{*}{Model}}&
			\multicolumn{2}{c}{DUTLF-V2}&\multicolumn{2}{c}{HFUT-LFSD}&\multicolumn{2}{c}{LFSD}\cr
			\cmidrule(lr){2-3} \cmidrule(lr){4-5} \cmidrule(lr){6-7}
			&$F^{w}_{\beta}\uparrow$&MAE$\downarrow$&$F^{w}_{\beta}\uparrow$&MAE$\downarrow$&$F^{w}_{\beta}\uparrow$&MAE$\downarrow$\cr
			\midrule
			Student&.721&.068&.651&.076&.647&.150\cr
			+MFD&.743&.064&.669&.073&.650&.149\cr
			+MFD+AFD&.749&.062&.667&.071&.678&.138\cr
			+MFD+AFD+SFD&\textcolor{red}{\bf.771}&\textcolor{red}{\bf.055}&\textcolor{red}{\bf.678}&\textcolor{red}{\bf.069}&\textcolor{red}{\bf.680}&\textcolor{red}{\bf.136}\cr
			\bottomrule
		\end{tabular}
	\end{threeparttable}
\end{table}
\begin{table}[t]
	\centering
	\setlength{\tabcolsep}{0.3mm}
	\begin{threeparttable}
		\caption{Ablation Analysis of the Proposed Distillation Schemes Based on Different Backbones (\textbf{\textcolor{red}{red bold:}} top-1 results)}
		\label{tab:ablation-student-backbone}
		\begin{tabular}{p{2.3cm}<{\centering}p{1cm}<{\centering}p{1cm}<{\centering}p{1cm}<{\centering}p{1cm}<{\centering}p{1cm}<{\centering}p{1cm}<{\centering}}
			\toprule
			\multicolumn{1}{c}{\multirow{2}{*}{Model}}&
			\multicolumn{2}{c}{DUTLF-V2}&\multicolumn{2}{c}{HFUT-LFSD}&\multicolumn{2}{c}{LFSD}\cr
			\cmidrule(lr){2-3} \cmidrule(lr){4-5} \cmidrule(lr){6-7}
			&$F^{w}_{\beta}\uparrow$&MAE$\downarrow$&$F^{w}_{\beta}\uparrow$&MAE$\downarrow$&$F^{w}_{\beta}\uparrow$&MAE$\downarrow$\cr
			\midrule
			ShuffleNet&.591&.093&.537&.099&.659&.142\cr
			+MFD&.624&.084&.553&.097&.661&.140\cr
			+MFD+AFD&.689&\textcolor{red}{\bf.073}&.621&\textcolor{red}{\bf.079}&.672&.139\cr
			+MFD+AFD+SFD&\textcolor{red}{\bf.691}&.074&\textcolor{red}{\bf.639}&.082&\textcolor{red}{\bf.701}&\textcolor{red}{\bf.130}\cr
			\midrule
			ResNet18&.637&.081&.565&.094&.668&.135\cr
			+MFD&.697&.065&.605&.082&.676&.137\cr
			+MFD+AFD&.730&.064&.655&.074&.678&.135\cr
			+MFD+AFD+SFD&\textcolor{red}{\bf.761}&\textcolor{red}{\bf.057}&\textcolor{red}{\bf.695}&\textcolor{red}{\bf.069}&\textcolor{red}{\bf.685}&\textcolor{red}{\bf.133}\cr
			\bottomrule
		\end{tabular}
	\end{threeparttable}
\end{table}

\subsubsection{Effect of MFSM.}
To provide evidence for the screening ability of MFSM, we visualized the saliency maps in different time steps as shown in Figure \ref{fig:MFSM}. We can observe that the attention module contributes more on locating salient object accurately in step 1 and 2, while the ConvLSTM contributes more on refining the details of salient object in step 3 and 4. The quantitative results are listed in Table \ref{tab:teacher-ablation}. Accumulative improvements are achieved as the time step increases.
These improvements are resonable
since the useful features are emphasized by the attention module
and spatial details are refined gradually by ConvLSTM.

\subsection{Knowledge Distillation Scheme Analysis}
To demonstrate the effectiveness of our proposed learning strategy, we conducted extensive ablations to explore the three distillation schemes, the multi-focusness distillation scheme, the attentive focusness distillation scheme and the screened focusness distillation scheme. The experiments are conducted on the student network (baseline VGG16) and all distillation schemes can independently produce fairly comparable results over the original student model.

\subsubsection{Multi-Focusness Distillation Scheme (MFD)}
\begin{figure}[!ht]
	\begin{center}
		\includegraphics[width=8.5cm]{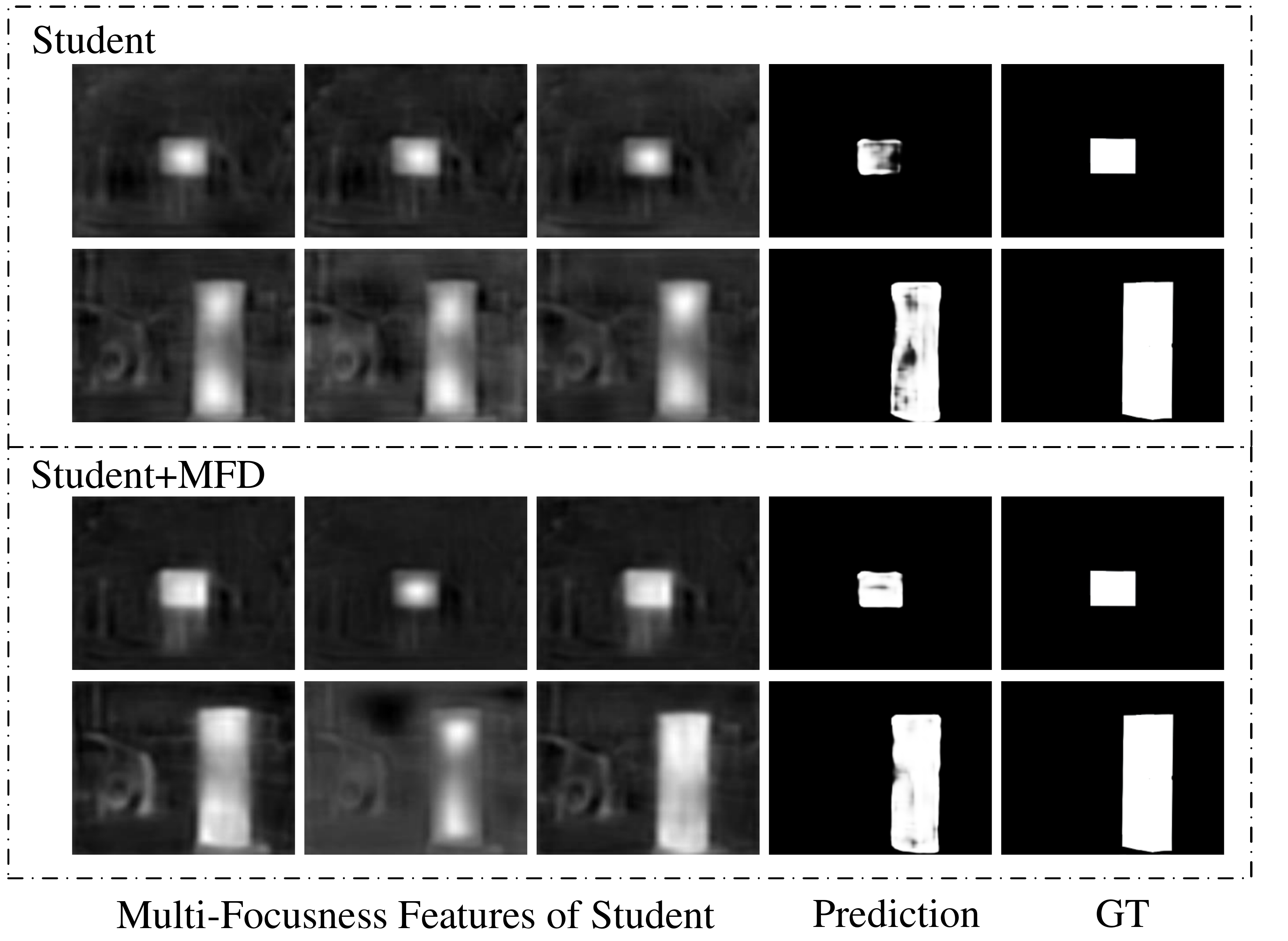}
	\end{center}
	\caption{Visualization of multi-focusness features of enabling and disabling MFD.}
	\label{fig:student-feature}
\end{figure}

The results in Table \ref{tab:ablation-student} show numerical improvements as we applied the MFD.
Specifically, the performance represents a 5.9\%, 3.9\% and 0.7\% increase on DUTLF-V2, HFUT-LFSD and LFSD in terms of MAE, respectively. 
Meanwhile, we can see from the visualization of multi-focusness features in Figure \ref{fig:student-feature} that the MFD focuses on transfering more effective multi-focusness features, such as features with finer boundary of salient object (row 3, column 1). Therefore, it produces more accurate prediction (row 3, column 4).

\begin{figure}[!ht]
	\begin{center}
		\includegraphics[width=8.5cm]{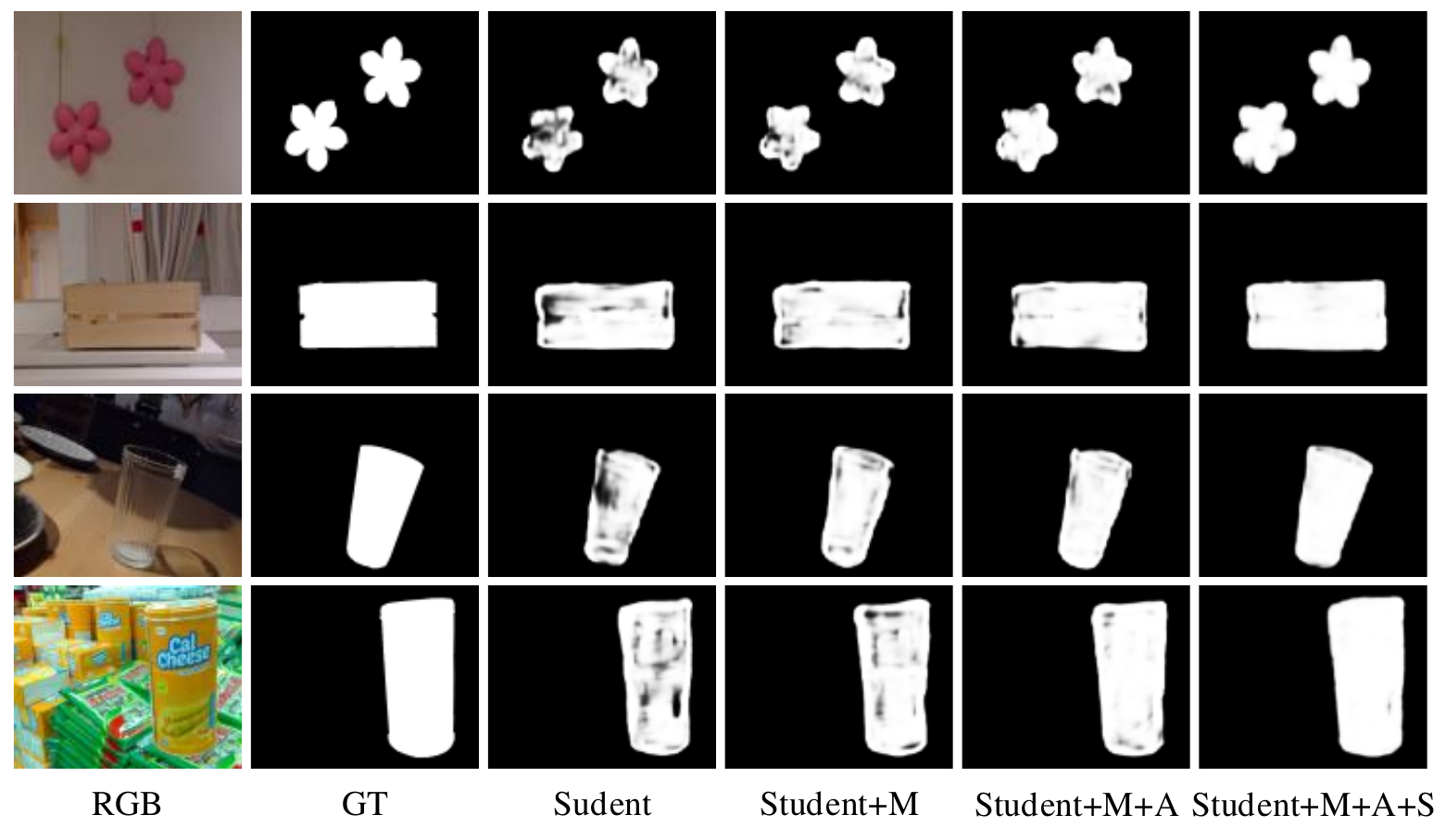}
	\end{center}
	\caption{Visual results of enabling and disabling different components of our distillation system. `Student+M', `Student+M+A' and `Student+M+A+S' refer to `Student+MFD', `Student+MFD+AFD' and `Student+MFD+AFD+SFD', respectively.}
	\label{fig:student-ablation-salmap}
\end{figure}
\subsubsection{Attentive Focusness Distillation Scheme (AFD)}
It can be seen that as we employed the AFD, the performance of the student network achieves accumulative improvements. Specifically, the performance represents a 3.1\%, 2.7\% and 7.4\% increase on DUTLF-V2, HFUT-LFSD and LFSD in terms of MAE, respectively.
The AFD transfers attentive knowledge that highlights the foreground, thus leads to more consistent saliency prediction (column 5 in Figure 8).

\subsubsection{Screened Focusness Distillation Scheme (SFD)}
As we further applied the SFD, the performance represents a 11.3\%, 2.8\% and 1.4\% increase on DUTLF-V2, HFUT-LFSD and LFSD in terms of MAE, respectively. 
We also visualized the results employing the SFD shown in Figure \ref{fig:student-ablation-salmap}. The SFD allows student network to learn complementarity of appearance and screened focusness information. Therefore, the student with SFD achieves a more detailed prediction.
When all distillation losses are employed, we obtained the sweet spot of the three. This validates the importance of focusness information transferring from the Focal steam to the RGB stream.

\begin{table*}[!t]
	\centering
	\setlength{\tabcolsep}{1mm}
	\begin{threeparttable}
		\caption{Quantitative Comparisons of E-measure, S-measure, Weighted F-measure, F-measure and MAE Scores on Three Light Field Datasets. The best three results are shown in \textbf{\textcolor{red}{red bold}},
			\textbf{\textcolor{blue}{bule bold}}, \textbf{\textcolor{green}{green bold}} fonts respectively
			($*$ represents conventional methods, - means no available results)}
		\label{tab:performance_comparison}
		\begin{tabular}{p{0.5cm}<{\centering}p{1cm}<{\centering}p{1.5cm}<{\centering}p{0.7cm}<{\centering}p{0.5cm}<{\centering}p{0.6cm}<{\centering}p{0.6cm}<{\centering}p{0.6cm}<{\centering}p{0.6cm}<{\centering}p{0.7cm}<{\centering}p{0.6cm}<{\centering}p{0.6cm}<{\centering}p{0.6cm}<{\centering}p{0.6cm}<{\centering}p{0.7cm}<{\centering}p{0.6cm}<{\centering}p{0.6cm}<{\centering}p{0.6cm}<{\centering}p{0.6cm}<{\centering}p{0.7cm}<{\centering}}
			\toprule
			\multicolumn{1}{c}{\multirow{2}{*}{Type}}&
			\multicolumn{1}{c}{\multirow{2}{*}{Methods}}&
			\multicolumn{1}{c}{\multirow{2}{*}{Years}}&
			\multicolumn{1}{c}{\multirow{2}{*}{\footnotesize{Size(M)$\downarrow$}}}&
			\multicolumn{1}{c}{\multirow{2}{*}{\footnotesize{FPS$\uparrow$}}}&
			\multicolumn{5}{c}{DUTLF-V2}&\multicolumn{5}{c}{HFUT-LFSD}&\multicolumn{5}{c}{LFSD}\cr
			\cmidrule(lr){6-10} \cmidrule(lr){11-15}\cmidrule(lr){16-20}
			&{}&{}&{}&{}&$E_s\uparrow$&$S_{\alpha}\uparrow$&$F^{w}_{\beta}\uparrow$&$F_{\beta}\uparrow$&\footnotesize MAE\small$\downarrow$&$E_s\uparrow$&$S_{\alpha}\uparrow$&$F^{w}_{\beta}\uparrow$&$F_{\beta}\uparrow$&\footnotesize MAE\small$\downarrow$&$E_s\uparrow$&$S_{\alpha}\uparrow$&$F^{w}_{\beta}\uparrow$&$F_{\beta}\uparrow$&\footnotesize MAE\small$\downarrow$\cr
			\midrule
			\multirow{2}{*}{}4D&Teacher&-&92.5&14&\textbf{\textcolor{red}{.924}}&\textbf{\textcolor{green}{.852}}&\textbf{\textcolor{red}{.792}}&\textbf{\textcolor{red}{.852}}&\textbf{\textcolor{blue}{.050}}&\textbf{\textcolor{red}{.858}}&.778&\textbf{\textcolor{blue}{.687}}&\textbf{\textcolor{red}{.753}}&\textbf{\textcolor{green}{.069}}&\textbf{\textcolor{blue}{.875}}&\textbf{\textcolor{green}{.827}}&\textbf{\textcolor{red}{.799}}&\textbf{\textcolor{red}{.852}}&\textbf{\textcolor{blue}{.087}}\cr
			2D&Student&-&\textbf{\textcolor{red}{47}}&\textbf{\textcolor{red}{105}}&\textbf{\textcolor{blue}{.909}}& .843&.771&\textbf{\textcolor{blue}{.813}}&.055&\textbf{\textcolor{green}{.850}}&.787&\textbf{\textcolor{green}{.678}}&.714&\textbf{\textcolor{green}{.069}}&.811&.737&.680&.738&.136\cr
			\midrule
			\multirow{5}{*}{4D}
			&LFNet&\footnotesize{TIP'20}&175.8&13&\textbf{\textcolor{green}{.907}}&\textbf{\textcolor{red}{.870}}&\textbf{\textcolor{blue}{.786}}&.803&\textbf{\textcolor{red}{.049}}&.840&\textbf{\textcolor{blue}{.800}}&\textbf{\textcolor{green}{.678}}&.701&\textbf{\textcolor{blue}{.065}}&.866&.812&.760&.814&.100\cr
			&DLFS&\footnotesize{IJCAI'19}&119&2&.839&.786&.641&.684&.087&.783&.741&.590&.615&.098&.806&.737&.657&.715&.147\cr
			&MoLF&\footnotesize{NIPS'19}&186.6&5&.866&.825&.709&.723&.065&.829&.789&.664&.678&.075&\textbf{\textcolor{red}{.886}}&\textbf{\textcolor{blue}{.830}}&\textbf{\textcolor{green}{.780}}&\textbf{\textcolor{green}{.819}}&\textbf{\textcolor{green}{.089}}\cr
			&LFS$^*$&\footnotesize{TPAMI'17}&-&-&-&-&-&-&-&.686&.579&.264&.430&.205&.771&.680&.479&.740&.208\cr
			&MCA$^*$&\footnotesize{TOOM'17}&-&-&-&-&-&-&-&-&-&-&-&-&.841&.749&-&.815&.150\cr
			&WSC$^*$&\footnotesize{CVPR'15}&-&-&-&-&-&-&-&-&-&-&-&-&.794&.706&.642&.706&.156\cr
			&DILF$^*$&\footnotesize{IJCAI'15}&-&-&.733&.648&.388&.504&.187&.736&.695&.458&.555&.131&.810&.755&.604&.728&.168\cr
			\midrule
			\multirow{7}{*}{3D}&DisenFusion&\footnotesize{TIP'20}&166&8&.836&.781&.636&.686&.093&.759&.734&.571&.605&.107&.831&.798&.724&.780&.115\cr
			
			&ATAFNet&\footnotesize
			{ACMMM'20}
			&291.5&21& .904&\textbf{\textcolor{blue}{.859}}&\textbf{\textcolor{green}{.775}}&\textbf{\textcolor{green}{.808}}&\textbf{\textcolor{green}{.051}}&\textbf{\textcolor{blue}{.856}}&\textbf{\textcolor{red}{.817}}&\textbf{\textcolor{red}{.712}}&\textbf{\textcolor{blue}{.736}}&\textbf{\textcolor{red}{.062}}&\textbf{\textcolor{green}{.868}}&.806&.761&.791&.103\cr
			&CPFP&\footnotesize{CVPR'19}&278&7&.843&.764&.629&.707&.075&.768&.701&.536&.594&.096&.669&.599&.465&.524&.186\cr
			&TANet&\footnotesize{TIP'19}&951.9&15&.849&.765&.609&.688&.090&.789&.744&.587&.638&.096&.849&.803&.727&.804&.112\cr
			&MMCI&\footnotesize{PR'19}&929.7&19&.829&.760&.545&.665&.109&.787&.741&.540&.645&.104&.848&.799&.685&.796&.128\cr
			&PCA&\footnotesize{CVPR'18}&533.6&15&.842&.762&.601&.683&.096&.782&.748&.598&.644&.095&.846&.807&.733&.801&.112\cr
			
			&CTMF&\footnotesize{Tcyb'17}&825.8&50&.836&.780&.573&.684&.104&.784&.752&.544&.620&.103&.856&.801&.710&.791&.119\cr
			
			\midrule
			\multirow{12}{*}{2D}
			&F$^3$Net&\footnotesize{AAAI'20}&102.5&92&.878&.841&.756&.803&.063&.810&.776&.673&.707&.094&.737&.758&.695&.742&.136\cr
			&SCRN&\footnotesize{ICCV'19}&96.6&29&.885&.847&.743&.790&.064&.823&\textbf{\textcolor{green}{.792}}&.676&\textbf{\textcolor{green}{.720}}&.093&.829&.796&.731&.763&.117\cr
			&EGNet&\footnotesize{ICCV'19}&412&21&.855&.821&.710&.746&.078&.794&.772&.634&.672&.094&.776&.784&.717&.762&.118\cr
			&CPD&\footnotesize{CVPR'19}&112&66&.885&.836&.754&.794&.061&.810&.764&.652&.689&.097&.865&\textbf{\textcolor{red}{.846}}&\textbf{\textcolor{blue}{.796}}&\textbf{\textcolor{blue}{.841}}&\textbf{\textcolor{red}{.083}}\cr
			&PoolNet&\footnotesize{CVPR'19}&278.5&32&.876&.832&.732&.774&.069&.802&.776&.652&.683&.092&.786&.800&.717&.769&.118\cr
			&PiCANet&\footnotesize{CVPR'18}&197.2&5&.850&.820&.685&.738&.086&.726&.781&.556&.618&.115&.780&.729&.621&.671&.158\cr
			
			&C2S&\footnotesize{ECCV'18}&158&30&.824&.787&.651&.691&.098&.786&.763&.630&.650&.111&.820&.806&.737&.749&.113\cr
			&R$^3$Net&\footnotesize{IJCAI'18}&225.3&3&.807&.724&.588&.643&.108&.728&.727&.566&.625&.151&.838&.789&.717&.781&.128\cr

			&DSS&\footnotesize{CVPR'17}&447.3&23&.828&.745&.597&.669&.101&.778&.715&.511&.626&.133&.749&.677&.570&.644&.190\cr
			\bottomrule
		\end{tabular}
	\end{threeparttable}
\end{table*}

\begin{figure*}[!ht]
	\begin{center}
		\includegraphics[width=1\linewidth]{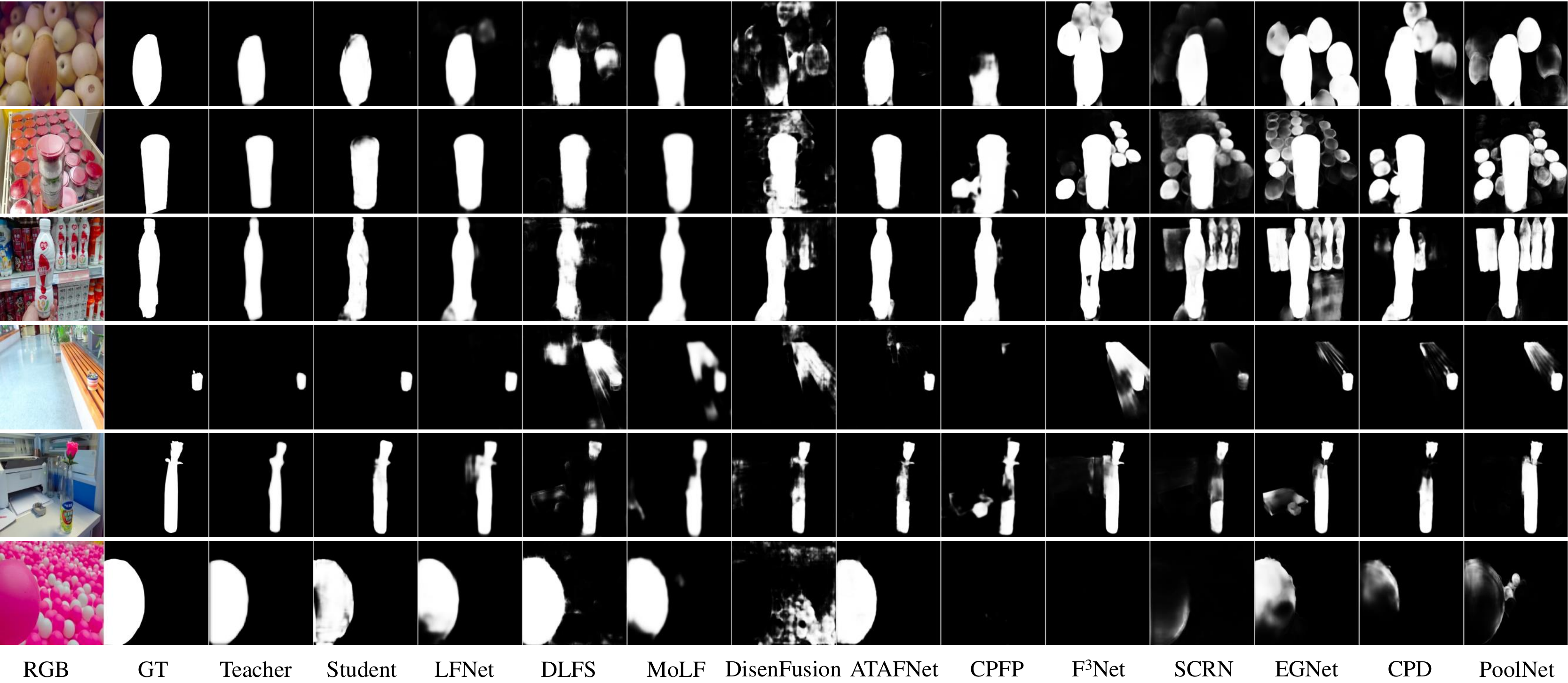}
	\end{center}
	\caption{Visual comparisons of our method with top-ranking CNNs-based methods in some challenging scenes.}
	\label{fig:visual-comparisions}
\end{figure*}
\subsection{Different Backbones}
In order to further verify the effect and generalization of the three distillation schemes, we extended our work by replacing the VGG16 with other two lightweight backbones.
As can be seen in Table \ref{tab:ablation-student-backbone}, the proposed distillation schemes contribute to accumulative performance improvements on all three datasets as the backbone is replaced by ShuffleNet or ResNet18. Specifically, in the setting with backbone ResNet18, our MFD, AFD and SFD reduce the MAE by 12.8\%, 9.8\% and 6.8\% on HFUT-LFSD, respectively. From the above observation, we can conclude that the three distillation schemes can effectively assist in predicting more accurate saliency results. Also, the improvement reflects high flexibility of the proposed distillation schemes that offer more benefits for a wide range of applications.

\subsection{{Comparison with State-of-the-arts}}
We compared our method with 23 state-of-the-art approaches including both deep-learning based methods and conventional methods (marked with `$*$'). There are \textbf{seven} 4D light field methods: LFNet \cite{zhang2020lfnet}, DLFS \cite{IJCAI19}, MoLF \cite{Zhang_2019_NeurIPS}, LFS$^*$ \cite{li2014saliency}, MCA$^*$ \cite{zhang2017saliency}, WSC$^*$ \cite{li2015weighted}, DILF$^*$ \cite{zhang2015saliency}; \textbf{seven} 3D RGBD methods: DisenFusion \cite{chen2020rgbd}, ATAFNet \cite{zhang2020feature}, CPFP \cite{zhao2019contrast}, TANet \cite{chen2019three}, MMCI \cite{chen2019multi}, PCA \cite{chen2018progressively}, CTMF \cite{han2017cnns}; and \textbf{nine} top-ranking 2D RGB methods: F$^3$Net \cite{wei2020f3net}, SCRN \cite{wu2019scrn}, EGNet \cite{zhao2019EGNet}, CPD \cite{wu2019cascaded}, PoolNet \cite{liu2019simple}, PiCANet \cite{liu2018picanet}, C2S \cite{li2018contour}, R$^3$Net \cite{deng2018r}, DSS \cite{hou2017deeply}. For a fair comparison, the results of the competing methods are generated by authorized codes or directly provided by authors.

\begin{table}[t]
	\centering
	\setlength{\tabcolsep}{0.7mm}
	\begin{threeparttable}
		\caption{Application of the Proposed Knowledge Distillation Schemes in Top-ranking RGB Saliency Models (\textbf{\textcolor{red}{red bold:}} top-1 results) }
		\label{tab:app}
		\begin{tabular}{p{3cm}<{\centering}p{1cm}<{\centering}p{0.7cm}<{\centering}p{0.7cm}<{\centering}p{0.7cm}<{\centering}p{0.7cm}<{\centering}}
			\toprule
			\multicolumn{1}{c}{\multirow{2}{*}{Methods}}&
			\multicolumn{1}{c}{\multirow{2}{*}{Size(M)$\downarrow$}}&
			\multicolumn{2}{c}{DUTLF-V2}&\multicolumn{2}{c}{HFUT-LFSD}\cr
			\cmidrule(lr){3-4} \cmidrule(lr){5-6}
			&{}&$F^{w}_{\beta}\uparrow$&\footnotesize MAE\small$\downarrow$&$F^{w}_{\beta}\uparrow$&\footnotesize MAE\small$\downarrow$\cr
			\midrule
			R$^3$Net-retrain&214.92&.757&.053&.656&.072\cr
			R$^3$Net+SFD&214.92&.775&.055&.680&.066\cr
			R$^3$Net+MFD+AFD+SFD&214.93&\textcolor{red}{\bf.793}&\textcolor{red}{\bf.051}&\textcolor{red}{\bf.716}&\textcolor{red}{\bf.062}\cr
			\midrule
			SCRN-retrain&96.71&.747&.055&.645&.078\cr
			SCRN+SFD&96.71&.768&.055&.677&.702\cr
			SCRN+MFD+AFD+SFD&96.72&\textcolor{red}{\bf.772}&\textcolor{red}{\bf.054}&\textcolor{red}{\bf.693}&\textcolor{red}{\bf.066}\cr
			\midrule
			CPD-retrain&111.56&.703&.071&.623&.088\cr
			CPD+SFD&111.56&.711&.068&.610&.084\cr
			CPD+MFD+AFD+SFD&111.57&\textcolor{red}{\bf.773}&\textcolor{red}{\bf.055}&\textcolor{red}{\bf.643}&\textcolor{red}{\bf.072}\cr
			\bottomrule
		\end{tabular}
	\end{threeparttable}
\end{table}

\subsubsection{The Teacher Network}
As can be seen from the quantitative results in Table \ref{tab:performance_comparison}, our proposed teacher network can largely outperform other advanced saliency models across all the datasets in terms of most evaluation metrics. It is worth mentioning that our significant advantages are achieved on the training set (3057 samples) three times smaller than the large RGB training set (10553 samples). We also show some visual results in Figure \ref{fig:visual-comparisions}. It can be easily seen that our teacher network can achieve more complete and accurate predictions in various challenging scenes, including similar foreground and background (row 1, 2 and 3), small or transparent salient object (row 4 and row 5), complex background (row 6).

\subsubsection{The Student Network}
\begin{figure}[!ht]
	\begin{center}
		\includegraphics[width=8cm]{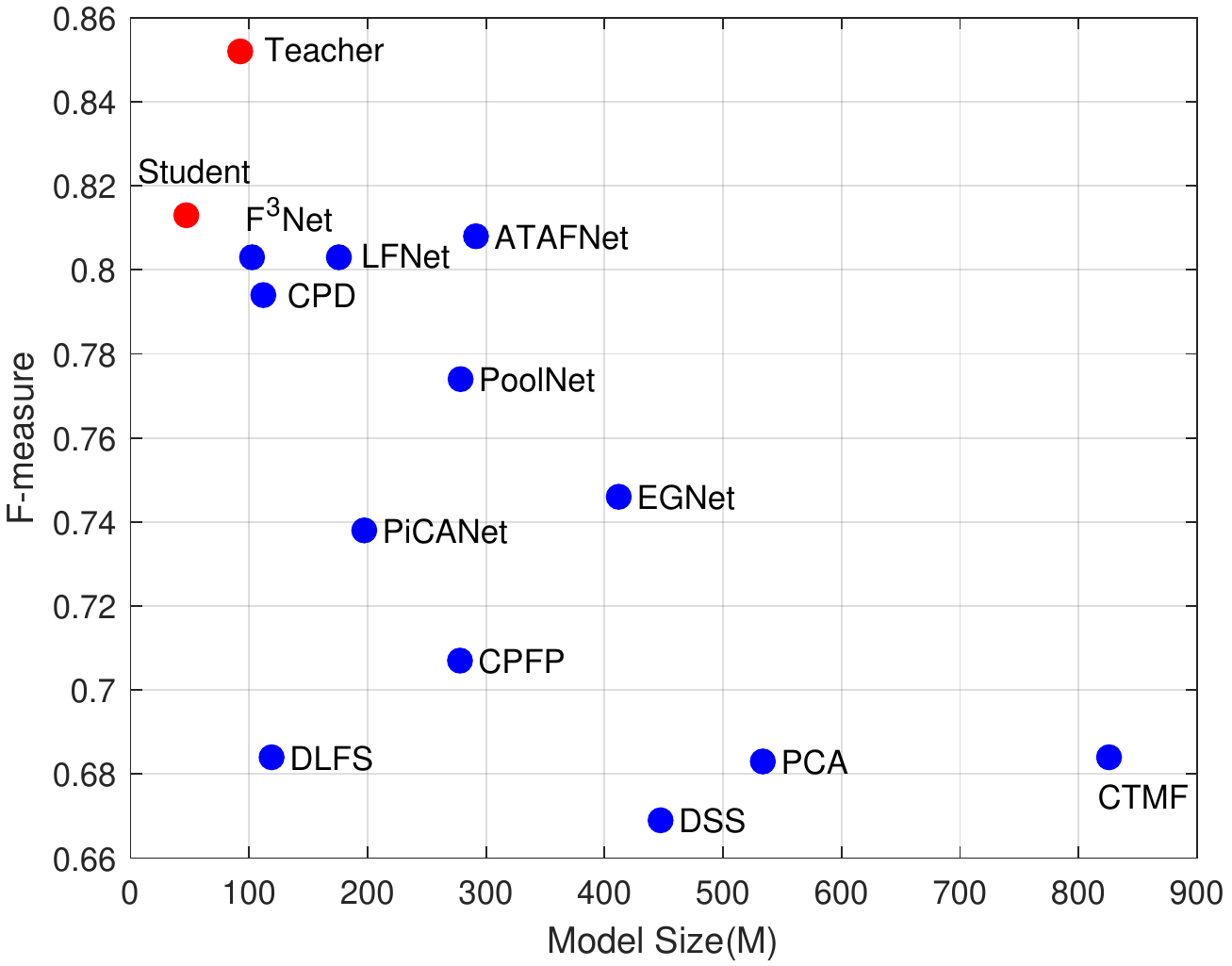}
	\end{center}
	\caption{Precision-Complexity scatterplot of the compared methods, abscissa represents model size, ordinate represents F-measure over the DUTLF-V2 dataset.}
	\label{fig:Size-F}
\end{figure}

\begin{figure}[!ht]
	\begin{center}
		\includegraphics[width=8cm]{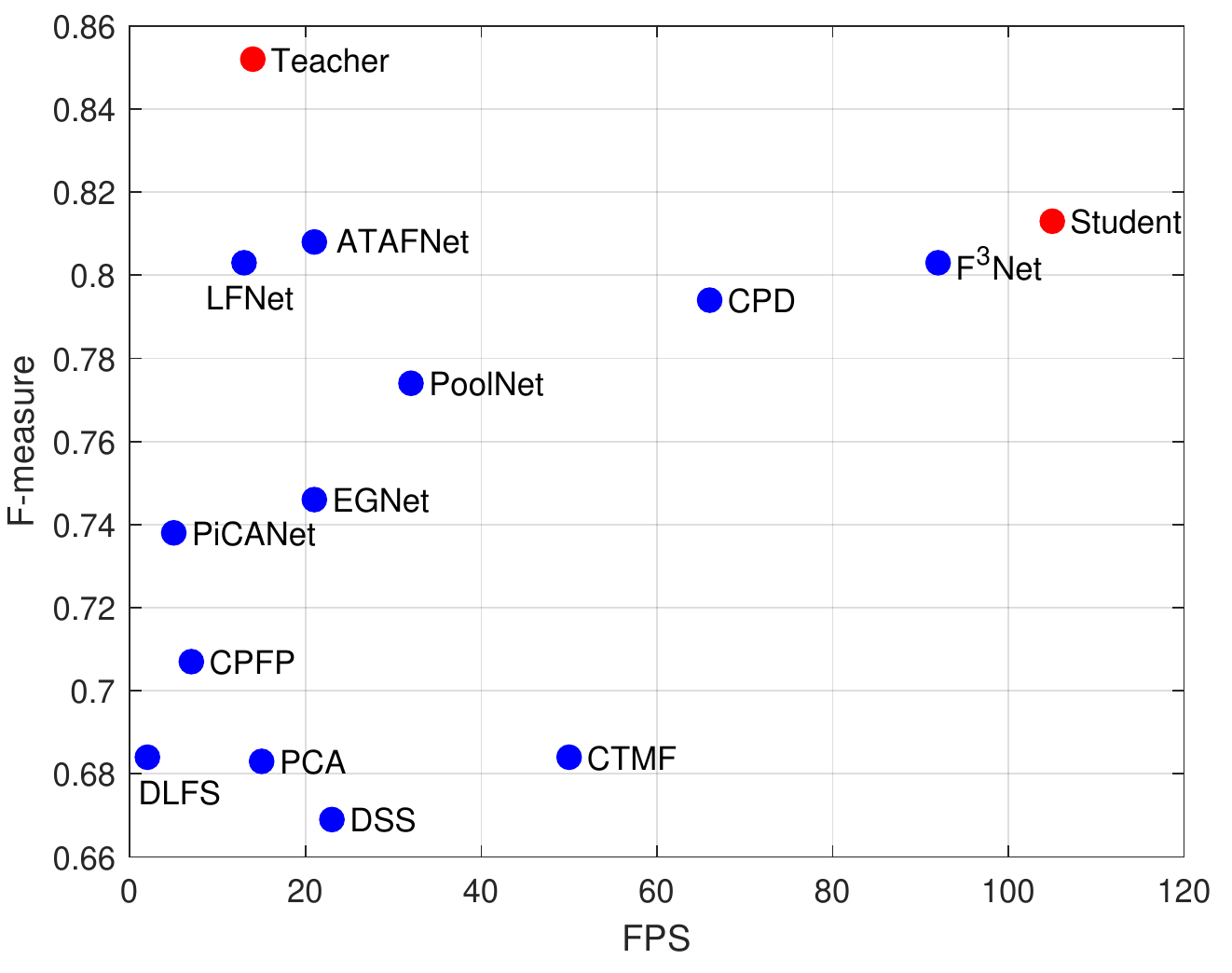}
	\end{center}
	\caption{Precision-Complexity scatterplot of the compared methods, abscissa represents FPS, ordinate represents F-measure over the DUTLF-V2 dataset.}
	\label{fig:FPS-F}
\end{figure}

Our proposed focusness knowledge distillation schemes can be seen as an effective replacement for the Focal stream, this leads to considerable results of the student network (VGG16), such as Top-2 F-measure on DUTLF-V2 dataset even with a single RGB input. Furthermore, we compared the average execution time and model size of our method with representative models. It is noted that the model size of our student network (VGG16) is only 47 MB and mean while FPS reaches up to 105. Compared to the currently best performing method ATAFNet, our student network tremedously minimizes the model size by 83\% and boosts the FPS by 5 times. 
The precision-complexity scatterplot in Figure \ref{fig:Size-F} and \ref{fig:FPS-F} also vividly demonstrate the superior performance of our student network.

\subsection{Application}
In this section, we applied the three distillation schemes to several top-ranking RGB saliency models, including R$^3$Net, SCRN and CPD. For a fair comparison, we retrained the original RGB models on the same training dataset of our method. The application is conducted via the following two settings with no or few extra parameters, both of which have obtained impressive performance gains. Specifically, for the no extra parameters setting, we trained the RGB models by directly applying the screened focusness distillation scheme which introduces no extra layers or parameters. For the few extra parameters setting, we adopted all three distillation schemes by adding a convolutional layer and a slice-wise attention module to impose the multi-focusness distillation scheme and attentive focusness distillation scheme, respectively. The added layers are lightweight and carry negligible computational burden, referred to Table \ref{tab:app}. Next, we will introduce details of the above two application patterns.

\begin{figure}[!ht]
	\begin{center}
		\includegraphics[width=8.5cm]{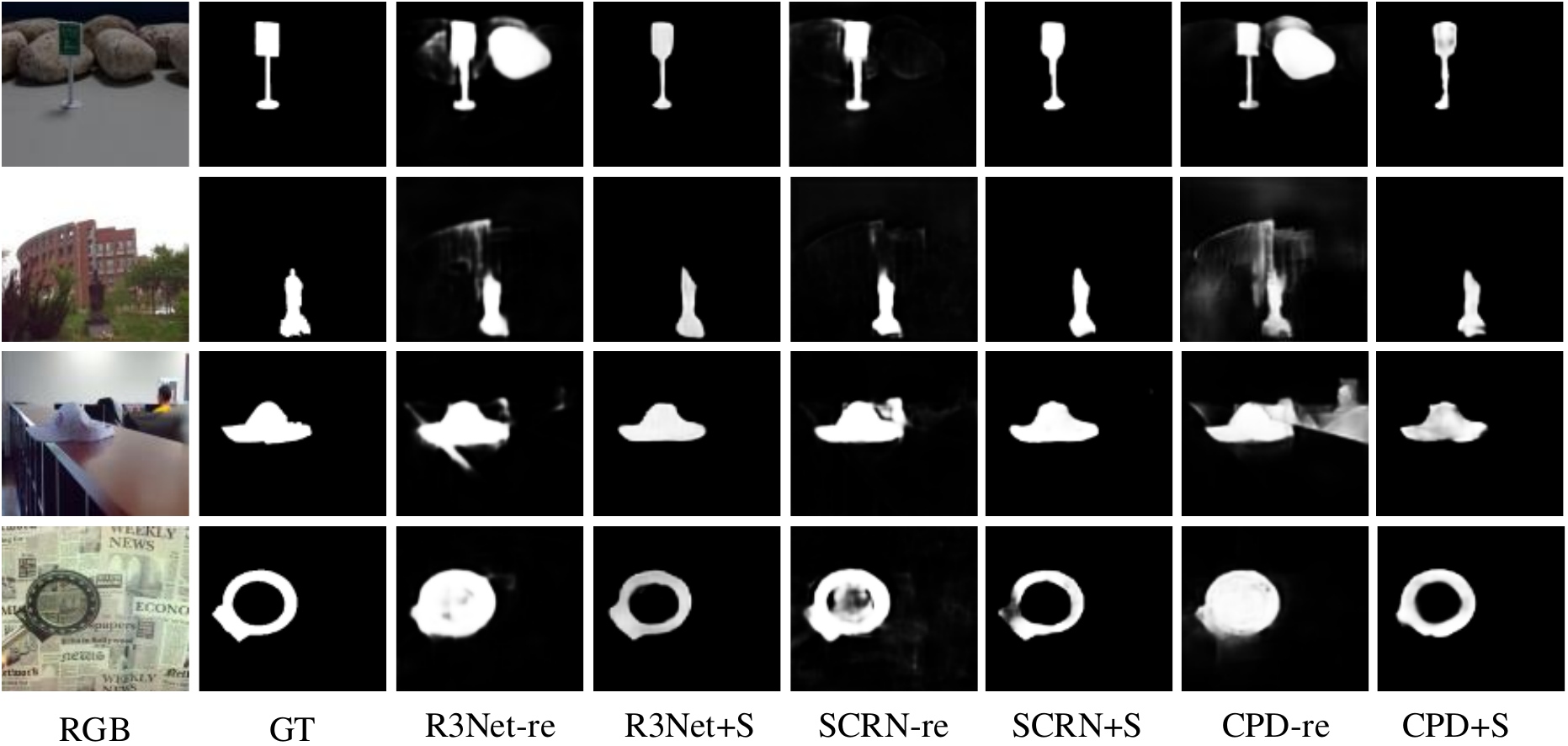}
	\end{center}
	\caption{Visualization of application with the no extra parameters setting. `R3Net-re', `R3Net+S' refer to `R3Net-retrain' and `R3Net+SFD', respectively.}
	\label{fig:app1}
\end{figure}

\begin{figure}[!ht]
	\begin{center}
		\includegraphics[width=8.5cm]{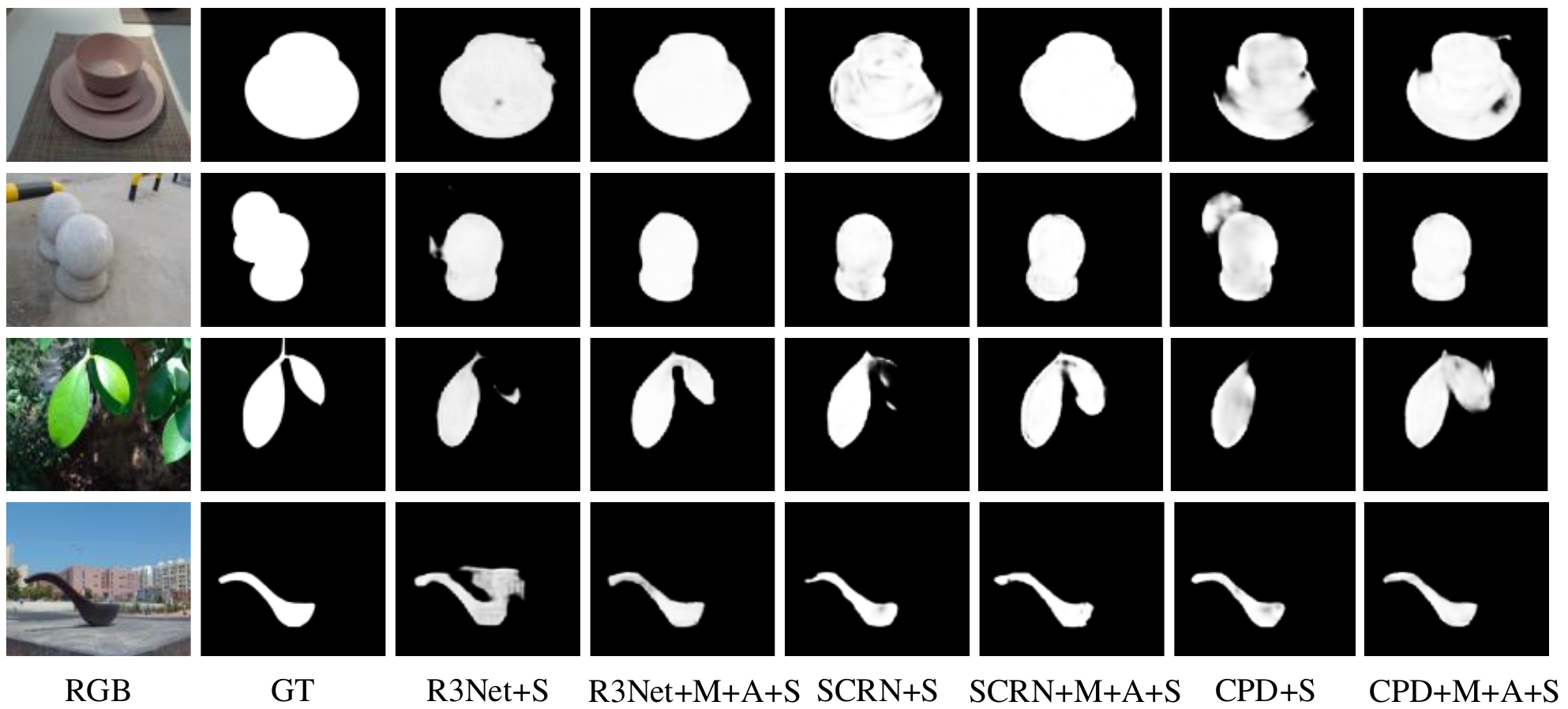}
	\end{center}
	\caption{Visualization of application with the few extra parameters setting. `R3Net+S', `R3Net+M+A+S' refer to `R3Net+SFD' and `R3Net+MFD+AFD+SFD', respectively.}
	\label{fig:app2}
\end{figure}

\subsubsection{No Extra Parameters Setting}
The first application is to embed the screened focusness distillation scheme in the existing models. In this way, there is no need to modify the structure of the original model, and no extra parameters are introduced. This ensures 
the flexibility and versatility of the application. It is noted that the models embedded with SFD are capable of learning complementary information from appearance and screened focusness information. As shown in Table \ref{tab:app}, R$^3$Net embedded with SFD (denoted as `R$^3$Net+SFD') has a consistent improvement compared with its original architecture (denoted as `R$^3$Net-retrain'), this application setting also improves the performance of SCRN and CPD. Besides, some visual comparisons of challenging examples are illustrated in Figure \ref{fig:app1}. We can note that after applying SFD, the saliency maps achieve superior consistence and improved robustness.

\subsubsection{Few Extra Parameters Setting}
We further applied all three distillation schemes to the advanced RGB models. This application setting slightly modified the original models and introduced few extra parameters, while achieving a large margin of improvement on performance. As shown in Table \ref{tab:app}, R$^3$Net adopted  MFD, AFD and SFD (denoted as `R$^3$Net+MFD+AFD+SFD') outperforms the original R$^3$Net (denoted as `R$^3$Net-retrain'), numerically reducing the MAE by 3.7\% and 13.8\% on DUTLF-V2 and HFUT-LFSD, respectively. The performance of other two models also benefits from our application. Meanwhile, the visual results in Figure \ref{fig:app2} illustrate that this application to RGB models helps them to predict more accurate saliency maps. It is also noted the comparison results of the two application settings demonstrate that the application of all three distillation schemes is more effective with few extra parameters.

\section{Discussion}
In our method, the proposed comprehensive and challenging DUTLF-V2 can assist in higher generalization for our models, and the three distillation schemes aim to ensure better absorption and integration of focusness knowledge for the student. This enables the student no longer need the focal slices but a single RGB image as input, and guarantees flexibility and productivity for mobile devices. But interestingly, we noticed that there is a certain performance gap between the student and the teacher. One possible reason we consider is that as the teacher network propagates forward and takes up a large percentage of GPU memory,  less computing resources are allocated to the student network and the batch size is set to 1, thus resulting in suboptimal performance and generalization of student network\cite{ioffe2015batch}.  We argue that this problem will be relieved as computing resources are updated.

\section{Conclusion}
In this paper, we constructed the DUTLF-V2 dataset, which contains a large scale of samples, diverse realistic scenes and a variety of data elements. Our proposed dataset provides a generic benchmark for the field of saliency detection.
Furthermore, we presented a novel asymmetrical two-stream network, which consists of Focal stream and RGB stream, to achieve efficiency and versatility for both desktop computers and mobile devices. 
On one hand, we consider the Focal stream as a teacher network, to learn to exploit focal slices and produce focusness knowledge tailored for student. Our proposed MFRM and MFSM recruit and screen useful saliency information effectively and enable the teacher network to achieve superior performance. 
On the other hand, we train the student network using single RGB input, to learn to replace focal slices relying on three tailor-made distillation schemes. The proposed distillation schemes allow the student to take advantage of multi-focusness information and ensure versatility and productivity for mobile devices.
Our evaluation shows the state-of-the-art performance of the teacher and the flexibility of the student.
Additionally, we demonstrate the applicability of the three proposed distillation schemes under a wide range of applications. Experimental results confirm the generalization and effectiveness of our distillation schemes.


%

\ifCLASSOPTIONcaptionsoff
  \newpage
\fi



%

{\small
\bibliographystyle{IEEEtran}
\bibliography{egbib}
}


%

\end{document}